\documentclass[11pt]{article}
\usepackage{quietbrand}
\usepackage{multirow}
\usepackage{float}
\usepackage[numbers,sort&compress]{natbib}
\usepackage{longtable}
\usepackage{tabularx}
\usepackage{placeins}
\usepackage{microtype}
\usepackage{newunicodechar}
\newunicodechar{≥}{\ensuremath{\geq}}
\newunicodechar{≤}{\ensuremath{\leq}}
\newunicodechar{×}{\ensuremath{\times}}

\PaperSetup{
  brand={PHILO AI}, brandcolor={22189C},
  abstractbackground=true, abstractframe=false,
  projectlinks=false, herofigure=true
}
\title{FlexVSR: Continuous-Scale Super-Resolution for Streaming Video Generation}
\author{\PaperAuthor{Ji Wang}{1}\PaperAuthorSep
  \PaperAuthor{Haonan An}{2}\PaperAuthorSep
  \PaperAuthor{Jiasheng Zhang}{1}\PaperAuthorSep
  \PaperAuthor{Renlong Chen}{1,*}}
\date{November 30, 2026}
\PaperAffiliation{\textsuperscript{1}Philo AI\quad
  \textsuperscript{2}City University of Hong Kong}
\PaperContact{\textsuperscript{*}Corresponding author: \href{mailto:reo@philoai.com}{\texttt{reo@philoai.com}}\quad
  First author: \href{mailto:jackwang@philoai.com}{\texttt{jackwang@philoai.com}}}
\PaperLogo{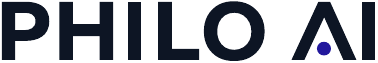}
\PaperFigure{\includegraphics[width=\linewidth]{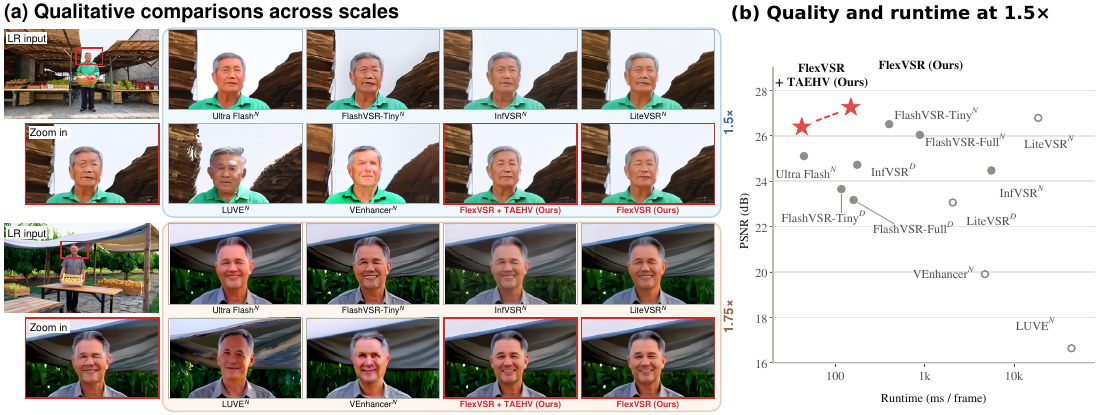}}{
(a)~Visual comparisons at FlexVSR's supervised ($1.5\times$) and unseen ($1.75\times$) scales.
(b)~PSNR vs.\ runtime at $1.5\times$ on one H100 GPU
(Sec.~\ref{sec:efficiency_analysis}; stars: FlexVSR with Wan~\cite{wan2025}/TAEHV~\cite{BoerBohan2025TAEHV}
decoding; filled circles: Streaming VSR methods; hollow circles: Offline VSR methods; $N$: official native-scale
inference followed by pixel-space downsampling; $D$: direct target-scale
inference).}{fig:teaser}

\begin{document}
\PaperMakeFrontmatter{Real-time streaming video generation has gained rapid momentum in live content creation and interactive media.
Deploying such systems in practice requires flexible scaling factors to
accommodate diverse source resolutions, target displays, and computational budgets. However, extending
fixed-scale generative video super-resolution (VSR) to varying target scales poses two challenges: downsampling from a larger fixed scale incurs redundant cost, while re-optimizing the diffusion transformer (DiT) is costly and alters the shared generative prior.
To address both, we propose \textbf{FlexVSR}, a decoupled framework for
continuous-scale streaming generative VSR with supervision at only a few scales. Our key insight is that naively resampling fixed-scale latents onto a new target grid induces scale-dependent input mismatch, which can be effectively addressed through parameter-efficient, scale-aware adaptation of the latent interface without updating the DiT.
To this end, FlexVSR introduces \textbf{ReLSA} (Refined Latent
Scale Adaptation), a lightweight latent adapter that constructs and refines target-grid latents before they enter the DiT, enabling a shared frozen VSR prior to generalize across continuous scales with only 0.274M trainable parameters and an adaptation cost of 7.3 H100 GPU-hours, while introducing no additional temporal dependencies and preserving the native causal streaming protocol.
Extensive experiments demonstrate competitive or superior
reconstruction and temporal quality, lower latency than most native-scale
streaming baselines and offline methods, and effective generalization to unseen continuous
scales, filling the continuous-scale gap in real-time streaming generative VSR. Project page: \url{https://philoai.com/research/flexvsr}.
}
\clearpage
\section{Introduction}
\label{sec:intro}

High-resolution video generation does not require performing every operation at the target resolution. Modern video generation pipelines widely adopt a coarse-to-fine paradigm: synthesizing global semantics and temporal dynamics at low resolution (LR), followed by latent upsampling and high-resolution (HR) generative refinement~\cite{Blattmann_2023_CVPR,he2024venhancer,Zhou_2024_CVPR,pmlr-v306-zhao26d,UltraFlash_2026,liu2026solrefiner}.
Within this paradigm, FlashVSR and InfVSR advance efficient streaming video
super-resolution (VSR)~\cite{Zhuang_2026_CVPR,pmlr-v306-zhang26dm}.
Yet in practical real-time deployment, scaling must flexibly bridge varying source inputs, target resolutions, and tight compute budgets.
Moderate, non-integer scales offer finer control over this quality--compute trade-off, which requires operating directly on the requested grid rather than restoring at a higher
resolution and then downsampling, since both DiT computation and costly pixel decoding~\cite{kang2026flashdecoder} scale steeply with output resolution. However, directly applying a fixed-scale model to naively resampled target-grid latents substantially degrades reconstruction quality, motivating lightweight scale adaptation of existing generative super-resolution (SR) priors.

For flexible scaling, image restorers have explored scale-conditioned
upsampling in Meta-SR~\cite{Hu_2019_CVPR} and continuous coordinate
querying in LIIF~\cite{Chen_2021_CVPR}.
In the video domain, discriminative methods such as BasicAVSR~\cite{Shang_2025_BasicAVSR} support online arbitrary-scale restoration yet lack generative priors for realistic detail synthesis. 
Meanwhile, generative approaches such as VEnhancer~\cite{he2024venhancer} and AVSR-Diff~\cite{youk2026avsr} combine diffusion priors with flexible spatiotemporal enhancement or continuous decoding, but rely on multi-step sampling and temporal lookahead that constrain real-time streaming deployment. 
Rather than designing and training an arbitrary-scale streaming diffusion model from scratch, we ask a more practical question: can an existing fixed-scale streaming generative VSR system be extended to continuous scales while strictly preserving its one-step causal inference protocol?

Reusing frozen generative priors via lightweight adapters is also well
established~\cite{Zhang_2023_ICCV,Mou_2024_AAAI}.
In restoration, methods such as StableSR~\cite{Wang_2024_StableSR},
DiffBIR~\cite{lin2024diffbir}, and PASD~\cite{yang2024pasd} preserve
pretrained synthesis models for image SR, while LiteVSR~\cite{pmlr-v306-cao26n} introduces a
state-aware adapter for a frozen video DiT.
Yet existing adapters do not address continuous-scale generalization of streaming SR priors, while re-optimizing a prepared SR DiT across scales incurs substantial training cost and modifies the shared reference-scale prior.
Crucially, keeping the SR DiT frozen does not exclude it from the optimization
loop: gradients can still backpropagate through the frozen restorer to
supervise an upstream scale adapter directly from the restored output.
\textbf{We therefore aim to achieve continuous-scale generalization from a few supervised scales while keeping the SR DiT frozen and streaming inference efficient.} 

We introduce \textbf{FlexVSR} (Fig.~\ref{fig:FlexVSR_Overview}), a streaming
generative video SR framework built for both flexible prior adaptation and flexible
output-scale selection. Targeting latent-cascade refinement in coarse-to-fine generation pipelines, FlexVSR decouples generative SR preparation from
subsequent scale adaptation, allowing a shared, frozen SR prior to 
accommodate diverse source resolutions and compute budgets across continuous target grids.
Reference-scale preparation first trains a fixed-scale latent upsampler and a one-step, chunk-causal SR DiT conditioned with historical latent reuse.

After preparation, the VAE, SR DiT, and upsampler feature extractor backbone remain
frozen, and we adapt the shared latent interface exclusively through
\textbf{ReLSA} (Refined Latent Scale Adaptation), a lightweight plug-in
adapter (Fig.~\ref{fig:FlexVSR_Overview}). ReLSA separates target-grid construction from
residual refinement: LIIF-style continuous querying~\cite{Chen_2021_CVPR}
first aggregates local LR features to predict a base latent, after which a
scale-conditioned residual branch incorporates aligned source features and
grid geometry to predict an additive correction for both DiT input routes.
Following a fast latent-regression warm-up, ReLSA is trained end-to-end at
only $1.5\times$ and $2\times$: its output is
noised as at inference and supervised in both latent and decoded-RGB spaces
through our frozen flow-matching
restorer~\cite{Lipman_2023_FlowMatching,Liu_2023_RectifiedFlow}, whose
one-step latent error is proportional to the velocity residual. 
At inference,
ReLSA runs frame-wise without temporal lookahead, enabling a single shared
prior to serve continuous scales while preserving single-step, chunk-causal
execution.

Experiments across multiple benchmarks demonstrate that FlexVSR achieves
competitive or superior reconstruction quality with lower latency than most
native-scale streaming and offline baselines. 
Cross-scale evaluations further verify effective generalization from the two supervised scales to unseen scale factors within the tested range.
Controlled analyses further show that closer target-latent proximity does not necessarily translate into better downstream restoration, and disentangle the gains of through-restorer backpropagation from decoded-space losses alone (Sec.~\ref{sec:prior_compatibility}).

Our contributions are threefold:
\begin{itemize}
    \item \textbf{A lightweight, scale-flexible streaming VSR paradigm.}
    FlexVSR decouples generative prior preparation from continuous-scale
    reconstruction via ReLSA, a 0.274M-parameter latent adapter
    combining continuous querying with scale-conditioned residual refinement.
    Trained end-to-end through the frozen SR DiT for prior compatibility in just 7.3 H100 GPU-hours,
    ReLSA preserves the shared prior and causal streaming protocol without
    temporal lookahead.
    
    \item \textbf{Generalization to unseen continuous scales.}
    Trained at only $1.5\times$ and $2\times$, a single shared adapter
    generalizes across unseen continuous scales, covering in-range
    interpolation and near-range extrapolation without scale-specific
    retraining for flexible inference-time scaling.

    \item \textbf{Efficient and transferable streaming VSR.}
    Direct target-grid execution avoids native-scale overhead while preserving strong quality, reaching ${\sim}24$~FPS with training-free TAEHV~\cite{BoerBohan2025TAEHV} decoding. Without DiT retraining, ReLSA also adapts diverse frozen priors, including Wan2.1-14B~\cite{wan2025}, MiniMax-H3~\cite{MiniMax_2026_H3}, and external Ultra Flash~\cite{UltraFlash_2026} (up to $30.9$~FPS; Sec.~\ref{sec:component_ablation}). We also introduce \textbf{VSR-17K} (17,377 training and 232 validation paired clips) to train prior preparation and adaptation, consistently boosting cross-scale quality (Sec.~\ref{sec:exp_setup}).
\end{itemize}

\section{Related Work}
\label{sec:related}

\subsection{Continuous-Scale SR and Latent Upsampling}
\label{sec:related_scale}
Continuous-scale SR aims to reconstruct arbitrary target grids with a single
model. In single-image SR, Meta-SR~\cite{Hu_2019_CVPR} predicts scale-conditioned
filters, ArbSR~\cite{Wang_2021_ICCV} adapts scale-specific networks via
plug-in modules, and implicit approaches such as LIIF~\cite{Chen_2021_CVPR},
LTE~\cite{Lee_2022_LTE}, and CiaoSR~\cite{Cao_2023_CiaoSR} query continuous
coordinates using local features, Fourier textures, or attention ensembles.
Extending to video, VideoINR~\cite{Chen_2022_CVPR} models continuous
space--time representations, SAVSR~\cite{Li_2024_SAVSR} combines
scale-adaptive convolution with spatiotemporal upsampling, and
LDIP~\cite{Bernasconi_2025_ICCV} couples long-distance propagation with a
reusable arbitrary-scale feature upsampler.
While these methods achieve continuous-scale spatial upsampling, their viability for streaming deployment depends on temporal aggregation design.
For instance, ST-AVSR~\cite{Shang_2024_STAVSR} attends to both past and
future frames, requiring future-frame lookahead, whereas
BasicAVSR~\cite{Shang_2025_BasicAVSR} 
explicitly provides online, limited-lookahead, and offline variants. While effective for pixel-space reconstruction, such discriminative designs focus on distortion-oriented regression rather than generative synthesis, whereas reusing a frozen latent restorer requires prior-compatible target-grid latents.
Meanwhile, generative approaches leverage diffusion priors through multi-step, offline temporal processing. VEnhancer~\cite{he2024venhancer} jointly enhances spatial and temporal resolution, using $\ge$15 sampling steps and overlapping chunks for long videos~\cite{VEnhancerCode_2024}. AVSR-Diff~\cite{youk2026avsr} denoises on an LR latent grid to limit computational growth with output scale, combining with 50 sampling steps and a bidirectional continuous decoder. LUVE~\cite{pmlr-v306-zhao26d} instead places latent upsampling at discrete scales between low-resolution generation and offline high-resolution content refinement. Thus, causal, continuous-scale generative VSR remains an open challenge.

\subsection{Streaming Generative Video Super-Resolution}
\label{sec:related_streaming}

Unlike offline restorers relying on bidirectional
propagation~\cite{Chan_2021_BasicVSR,Chan_2022_BasicVSRPP} or multi-step
diffusion~\cite{Zhou_2024_CVPR}, streaming generative VSR must jointly
constrain temporal lookahead and denoising steps.
Through one-step distillation, FlashVSR~\cite{Zhuang_2026_CVPR} combines
locality-constrained sparse attention with lightweight conditional decoding,
while InfVSR~\cite{pmlr-v306-zhang26dm} pairs a causal DiT with rolling KV
caching.
For latent generation cascades, SoL-Refiner~\cite{liu2026solrefiner} achieves
fast one-step high-resolution refinement, while Ultra
Flash~\cite{UltraFlash_2026} further enables real-time streaming via causal
latent upsampling and cascaded one-step refinement.
Since these models are tailored to fixed target scaling factors, extending them to arbitrary target grids requires either extra native-scale computation and decoding before downsampling or direct frame or latent resizing with reduced reconstruction quality.

\subsection{Efficient Adaptation of Generative Priors}
\label{sec:related_adaptation}

Diffusion models~\cite{Ho_2020_DDPM,Rombach_2022_LDM}, diffusion
transformers~\cite{Peebles_2023_DiT}, and flow
matching~\cite{Lipman_2023_FlowMatching,Liu_2023_RectifiedFlow} provide
powerful, reusable generative priors.
To adapt frozen backbones, parameter-efficient tuning such as
LoRA~\cite{Hu_2022_LoRA} and side-branch controllers including
ControlNet~\cite{Zhang_2023_ICCV} and T2I-Adapter~\cite{Mou_2024_AAAI} train
lightweight modules without updating base weights.
In super-resolution, StableSR~\cite{Wang_2024_StableSR} uses a time-aware
encoder for image SR, while LiteVSR~\cite{pmlr-v306-cao26n} conditions a frozen
video DiT on low-quality inputs and intermediate denoising states.
Unlike fixed-scale adapters that inject internal conditioning into offline DiTs, FlexVSR unifies continuous upsampling and streaming restoration by adapting only the upstream latent interface, generalizing a frozen one-step causal DiT across target grids from sparse scale supervision while preserving its weights and causal protocol.
\section{Method}
\label{sec:method}
\begin{figure}[tbp]
    \centering
    \includegraphics[width=\textwidth]{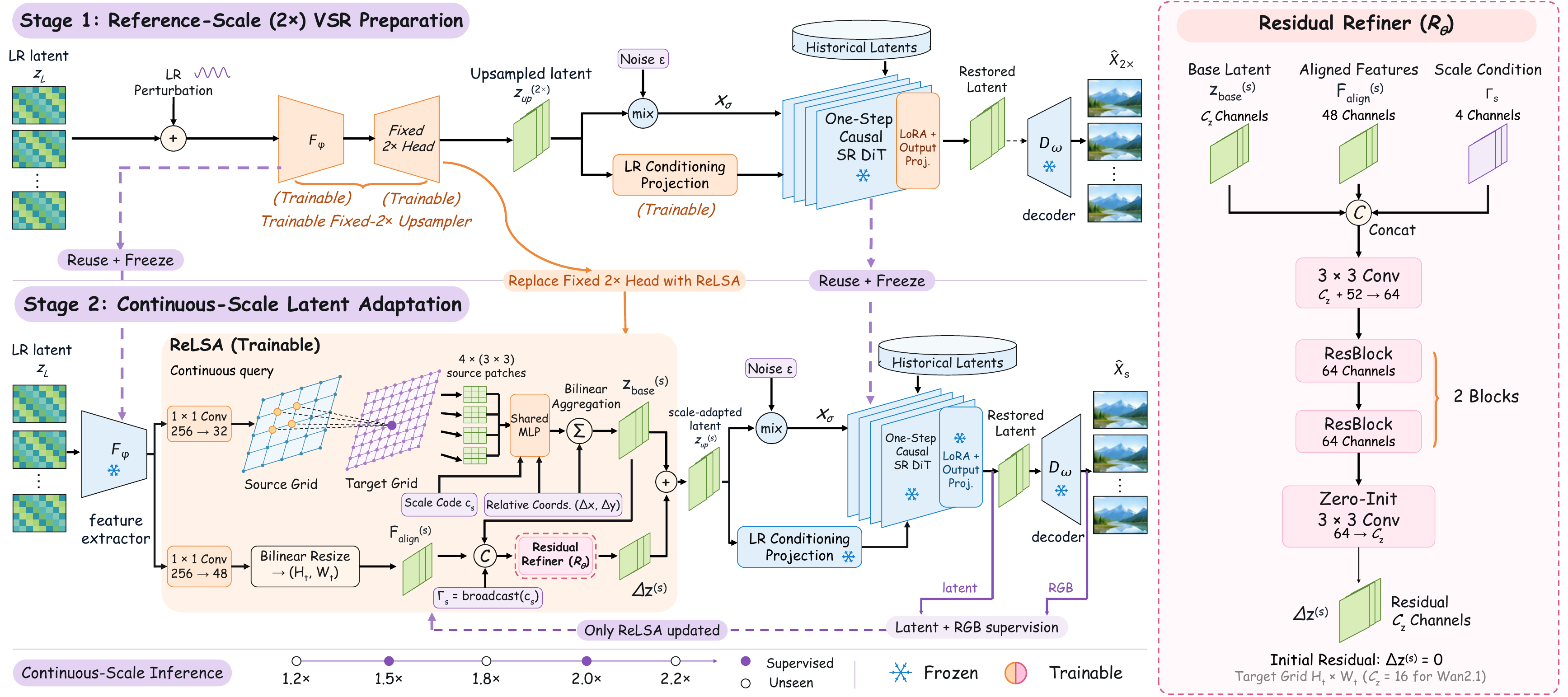}
\caption{\textbf{Overview of FlexVSR.} \textbf{Left:} Stage~1 prepares a $2\times$ streaming SR prior via latent velocity regression (dashed arrow before $D_\omega$ denotes inference-only decoding). Stage~2 freezes the restorer and feature extractor $F_\varphi$, replacing the $2\times$ head with \textbf{ReLSA} to adapt latents across scales under discrete $1.5\times$/$2\times$ supervision (latent warm-up followed by end-to-end latent+RGB training). \textbf{Right:} Architecture of the target-grid \textbf{Residual Refiner ($R_\theta$)}, zero-initialized at the output so $\Delta z^{(s)}=0$ initially.}
    \label{fig:FlexVSR_Overview}
\end{figure}

To enable continuous-scale streaming VSR without degrading restoration quality through direct latent resampling, FlexVSR operates in two decoupled stages (Fig.~\ref{fig:FlexVSR_Overview}): one-time preparation of a reference-scale streaming SR prior, followed by lightweight upstream latent adaptation via ReLSA across continuous scales.

\subsection{Preliminaries and Problem Definition}
\label{sec:adapter}

\noindent\textbf{Preliminary: Flow-matching streaming restoration.}
Flow matching~\cite{Lipman_2023_FlowMatching,Liu_2023_RectifiedFlow} defines a
linear interpolation path between a clean target latent $Y$ and Gaussian noise
$\epsilon\sim\mathcal N(0,I)$ at noise level $\sigma\in[0,1]$, training a
velocity predictor $v_\psi$ toward target velocity $v^*$:
\begin{equation}
x_\sigma = (1-\sigma)Y + \sigma\epsilon, \qquad
v^* = (Y - x_\sigma)/\sigma.
\label{eq:fm_prelim}
\end{equation}
For real-time VSR, streaming designs pair a chunk-causal
DiT~\cite{Peebles_2023_DiT} with historical latent context
$\mathcal H$~\cite{UltraFlash_2026,pmlr-v306-zhang26dm}, initializing $x_\sigma$ from a
coarse upsampled latent $z_{\mathrm{up}}$ (in place of $Y$ at small $\sigma$)
to restore $\widehat Y = x_\sigma + \sigma\,v_\psi(\cdot)$ in one Euler step.

\noindent\textbf{Problem definition: Continuous-scale latent adaptation.}
Given a streaming LR video latent
$Z_L\in\mathbb R^{T\times H\times W\times C_z}$ and continuous scale $s$ for
target grid $(H_t,W_t)$, we aim to causally reconstruct $\widehat X_s$ without
future-frame lookahead or backbone retraining.
Keeping a prepared feature extractor $F_\phi$, one-step streaming SR DiT
$\mathcal S_\psi$, and VAE decoder $D_\omega$ frozen, FlexVSR formulates scale
generalization as learning an upstream latent adapter $A_\theta$
(ReLSA):
\begin{equation}
\begin{aligned}
F &= F_\phi(Z_L),\\
z_{\mathrm{up}}^{(s)} &= A_\theta(F;H_t,W_t),\\
\widehat X_s &= D_\omega\!\left(\mathcal S_\psi
(z_{\mathrm{up}}^{(s)};p,\mathcal H,\epsilon)\right),
\end{aligned}
\label{eq:relsa_pipeline}
\end{equation}
where $p$ is the inherited text condition.
Within $\mathcal S_\psi$, $z_{\mathrm{up}}^{(s)}$ instantiates both the noisy
starting state $x_\sigma$ (substituting for $Y$ in Eq.~\ref{eq:fm_prelim}) and
the patch-packed ($\Pi$) content condition $c_L$ via conditional projection
$P_{\mathrm{LR},\psi}$ (\texttt{proj\_lr}):
\begin{equation}
x_\sigma = (1-\sigma)z_{\mathrm{up}}^{(s)}+\sigma\epsilon, \quad
c_L = P_{\mathrm{LR},\psi}\!\left(\Pi(z_{\mathrm{up}}^{(s)})\right).
\label{eq:sr_state_condition}
\end{equation}

\noindent\textbf{Stage 1: Reference-scale prior preparation.}
On VSR-17K (Sec.~\ref{sec:exp_setup}), we first train a fixed-$2\times$ latent
upsampler ($F_\phi$ and a $2\times$ head outputting $z_{\mathrm{up}}^{(2\times)}$)
to prepare $\mathcal S_\psi$ based on Wan2.1-1.3B~\cite{wan2025}.
With the VAE~\cite{Kingma_2014_VAE,wan2025} frozen, we optimize
LoRA~\cite{Hu_2022_LoRA} weights alongside $P_{\mathrm{LR},\psi}$ and output
projections via velocity regression (Eq.~\ref{eq:fm_prelim}).

\noindent\textbf{Stage 2: Latent-scale adaptation.}
We bypass the original $2\times$ upsampling and output layers, keeping
$F_\phi$, $\mathcal S_\psi$ (including all Stage-1 adaptations and
$P_{\mathrm{LR},\psi}$), and $D_\omega$ frozen.
ReLSA $A_\theta$ constructs target-grid latents $z_{\mathrm{up}}^{(s)}$ from
$F$ and is trained end-to-end through the frozen $\mathcal S_\psi$
(Sec.~\ref{sec:Decoder}) on REDS~\cite{nah2019ntire} and VSR-17K
(Sec.~\ref{sec:exp_setup}).

\subsection{Target-Grid Latent Refinement}
\label{sec:refine}

Operating frame-wise, ReLSA combines continuous base-latent construction with
feature-con\-di\-tioned residual refinement (Fig.~\ref{fig:FlexVSR_Overview}, Stage~2 and right). Given source
size $(H,W)$ and target size $(H_t,W_t)$, both branches condition on the
four-dimensional scale code
\begin{equation}
\mathbf c_s=\left[
\frac{H}{H_t},\frac{W}{W_t},
\frac{H_t}{H},\frac{W_t}{W}
\right].
\label{eq:relsa_scale_code}
\end{equation}

\noindent\textbf{Base latent construction.}
\label{sec:Continuous}
Following local implicit representations~\cite{Chen_2021_CVPR}, a $1\times1$
convolution projects the frozen 256-channel source features $F$ to 32 channels.
For each target query position, a shared MLP predicts a latent vector from each
of the four neighboring $3\times3$ source patches, conditioned on relative
coordinates and $\mathbf c_s$. Bilinear-weighted aggregation then yields the
base latent $z_{\mathrm{base}}^{(s)}\in\mathbb R^{H_t\times W_t\times C_z}$.

\noindent\textbf{Learned residual refinement.}
Architecturally, the continuous-query branch predicts each target position independently from local source features, relative coordinates, and $c_s$, without explicit spatial refinement across neighboring target positions; this coordinate-wise construction can leave local, scale-dependent inconsistencies in the assembled target-grid latent. 
To complement it with spatial context, a parallel learned $1\times 1$ projection $P_{48,\theta}$ maps $F$ to 48 channels, followed by bilinear resizing to align with the target grid. Let
$\Gamma_s\in\mathbb R^{H_t\times W_t\times 4}$ denote $\mathbf c_s$ broadcast
over the target grid; the refiner $R_\theta$ computes
\begin{equation}
\begin{aligned}
F_{\mathrm{align}}^{(s)}
&=\operatorname{Resize}_{H_t,W_t}\!\left(P_{48,\theta}(F)\right),\\
\Delta z^{(s)}
&=R_\theta\!\left([z_{\mathrm{base}}^{(s)},
F_{\mathrm{align}}^{(s)},\Gamma_s]\right),\\
z_{\mathrm{up}}^{(s)}
&=z_{\mathrm{base}}^{(s)}+\Delta z^{(s)},
\end{aligned}
\label{eq:relsa_refinement}
\end{equation}
where $[\cdot]$ denotes channel-wise concatenation of $C_z+48+4=C_z+52$
channels ($68$ for Wan2.1-1.3B~\cite{wan2025} with $C_z=16$). Here, $R_\theta$
uses a $3\times3$ input convolution (hidden width 64), two residual
blocks~\cite{He_2016_ResNet}, and a zero-initialized $3\times3$ output
convolution~\cite{Zhang_2023_ICCV} so $z_{\mathrm{up}}^{(s)}$ initially
equals $z_{\mathrm{base}}^{(s)}$.

\subsection{Training through the Frozen Restorer and Streaming Inference}
\label{sec:Decoder}
\label{sec:Streaming}

\noindent\textbf{End-to-end training via velocity residuals.}
We encode matched RGB clips at the LR and target resolutions with the frozen
Wan VAE~\cite{wan2025}, yielding $Z_L$ and native-scale target latents $Y_s$.
Feeding the adapted latent $U_s = z_{\mathrm{up}}^{(s)}$ into
Eq.~\ref{eq:sr_state_condition} at $\sigma = 0.2$ yields $(x_\sigma, c_L)$ for
the frozen one-step update
$\widehat Y_s = x_\sigma + \sigma\, v_\psi(x_\sigma,\sigma,c_L,p,\mathcal H)$.
With the Stage-1 target velocity $v^* = (Y_s - x_\sigma)/\sigma$, the post-update restoration error equals the $\sigma$-scaled velocity residual at the adapter output:
\begin{equation}
\widehat Y_s - Y_s = \sigma\!\left(v_\psi(x_\sigma,\sigma,c_L,p,\mathcal H) - v^*\right).
\label{eq:restore}
\end{equation}

Keeping $\mathcal{S}_{\psi}$ and $D_{\omega}$ frozen after a fast
latent-Charbonnier warm-up, we optimize ReLSA to shape both DiT input routes
(Eq.~\ref{eq:sr_state_condition}) and minimize this post-update residual using a latent anchor on $U_s$,
Charbonnier~\cite{Charbonnier_1997_TIP} and temporal-difference losses in both latent
and RGB spaces, LPIPS~\cite{Zhang_2018_CVPR}, and DISTS~\cite{ding2022dists}. To bound memory and
runtime, RGB losses decode only two adjacent latent frames from a $32\times32$
latent crop into four $224\times224$ frames after border removal. Each step restores four-frame latent chunks with no history (25\%) or two history frames (75\%, sampling detached predictions with probability 0.75 and ground truth otherwise) to mitigate streaming drift (full training procedure and loss weights in Appendix~\ref{sec:supp_training}).

\noindent\textbf{Streaming inference.}
At inference, ReLSA processes latent frames independently without future-frame
lookahead or temporal caching, while the frozen SR DiT retains its
chunk-causal execution and historical latent context~\cite{pmlr-v306-zhang26dm,UltraFlash_2026}.
Constructing the target $(H_t,W_t)$ latent grid prior to restoration scales
DiT token computation directly with the requested resolution~\cite{Peebles_2023_DiT}.

\section{Experiments}
\label{sec:experiments}

\subsection{Experimental Setup}
\label{sec:exp_setup}

\noindent\textbf{Model and training.}
We instantiate FlexVSR with a Wan2.1-1.3B~\cite{wan2025} restoration prior.
With the feature extractor, SR DiT, and Wan VAE frozen, scale adaptation
updates only the 273,584-parameter ReLSA adapter under equal $1.5\times$ and
$2\times$ supervision. Training uses 40k latent-regression warm-up iterations followed by 1k end-to-end (E2E) iterations through the frozen restorer (Sec.~\ref{sec:Decoder}).

\noindent\textbf{Training data.}
Stage~1 prepares the base prior on \textbf{VSR-17K}, and Stage~2 trains ReLSA
on all 20,521 clips from REDS~\cite{nah2019ntire} (3,144, excluding
REDS4) and VSR-17K (17,377 training, with 232 source-disjoint validation), plus a REDS-only control (Sec.~\ref{sec:component_ablation}).
Curated from quality-filtered public videos ($\ge$10 Mbps, 96.6\% 4K) and screened for test-set overlap via sampled-frame perceptual hashes, VSR-17K offers paired 21-frame clips independently resized in RGB and VAE-encoded per target scale (see Appendix~\ref{sec:supp_vsr17k}).

\noindent\textbf{Evaluation.}
We evaluate main comparisons at $1.5\times$ and $2\times$ on
YouHQ40~\cite{Zhou_2024_CVPR} (37 landscape clips at $1056\times1920$ and 3
square clips at $1056\times1056$, 31--33 frames each) and Inter4K-40, a fixed
40-clip subset of Inter4K~\cite{stergiou2023adapool} (81 frames each). Scale
generalization and ablation studies are conducted on YouHQ40. All quality
metrics are computed on full-frame RGB outputs against ground-truth HR frames
at matched spatial and temporal resolutions.

\noindent\textbf{Compared systems.}
Focusing on adapting generative VSR priors, we primarily compare FlexVSR with representative fixed- and multi-scale generative VSR systems across streaming and offline settings.
Most compared systems use a Wan2.1-1.3B~\cite{wan2025} DiT, including
FlashVSR~\cite{Zhuang_2026_CVPR} (with Tiny \text{-T} and
Full \text{-F} decoders), whereas LiteVSR~\cite{pmlr-v306-cao26n} uses Wan2.2-5B~\cite{wan2025} (official 5-step setting) and VEnhancer uses an I2VGen-XL 3D U-Net
with SD2.1 spatial layers (15-step fast mode, trained across
$1\times$--$8\times$)~\cite{he2024venhancer}. Among latent baselines, LUVE
retains VLUer and HCR ($1.5\times/2\times/3\times$)~\cite{pmlr-v306-zhao26d},
and Ultra~Flash provides separate $2\times$ and $3\times$
models~\cite{UltraFlash_2026}. 
Since upstream T2V generations lack paired multi-scale ground truth for reference-based evaluation, we follow standard cascade benchmarks by keeping each HR reference fixed and synthesizing scale-specific LR frames via bicubic downsampling (while showing qualitative results from actual T2V-generated latents in Fig.~\ref{fig:teaser}, Sec.~\ref{sec:qualitative_generated}, and Appendix~\ref{sec:supp_qualitative}). All systems receive identical LR frames at each scale via their respective RGB or latent interfaces, and we evaluate released checkpoints without modifying model weights.
Matching Fig.~\ref{fig:teaser}, we distinguish two evaluation routes for fixed-scale
baselines: $N$ (official native inference settings) restores at the
model's native scale and downsamples the output to the target size (for the
$4\times$ models LiteVSR, FlashVSR, and InfVSR, route $N$ processes
$7.1\times/4\times$ the target pixel count at $1.5\times/2\times$), whereas
$D$ (direct target-scale inference) resizes the input to restore
directly on the target grid.

\noindent\textbf{Metrics.}
We evaluate spatial fidelity and perceptual quality using RGB PSNR and
LPIPS~\cite{Zhang_2018_CVPR}, and temporal consistency via
tLPIPS~\cite{chu2020tecogan}, which measures the absolute difference between
output and ground-truth adjacent-frame LPIPS. Metrics are averaged per clip and
then across clips.

\subsection{Comparison with Generative VSR Systems}
\label{sec:system_comparison}
\begin{table}[!htb]
\centering
\setlength{\aboverulesep}{0pt}\setlength{\belowrulesep}{0pt}
\caption{\textbf{Quantitative comparison with generative VSR systems.}
Best and second-best scores
are \textbf{bold} and \underline{underlined}.}
\label{tab:quality_comparison}
\label{tab:method_context}
\fontsize{9}{11.0}\selectfont
\setlength{\tabcolsep}{2.0pt}

\renewcommand{\arraystretch}{1.20}
\newcommand{\FlexVSRctx}[1]{\hspace{5pt}\begin{tabular}[c]{@{}l@{}}#1\end{tabular}}
\newcommand{\FlexVSRtype}[2]{\raisebox{#1}[0pt][0pt]{\itshape\begin{tabular}[c]{@{}c@{}}#2\end{tabular}}}
\sbox0{%
\begin{tabular}{c@{\hspace{5pt}}|c@{\hspace{5pt}}|@{}l@{\extracolsep{0pt}}*{7}{c}@{}}
\toprule
\multirow{2}{*}{Dataset} &
\multirow{2}{*}{Type} &
\multicolumn{1}{@{}c}{\multirow{2}{*}{Method (venue)}} &
\multicolumn{3}{c}{$1.5\times$} & \multicolumn{3}{c}{$2\times$} &
\multirow{2}{*}{Avg.\ Rank\,$\downarrow$} \\
\cmidrule(lr){4-6}\cmidrule(lr){7-9}
{} & {} & {} &
PSNR\,$\uparrow$ & LPIPS\,$\downarrow$ & tLPIPS\,$\downarrow$ &
PSNR\,$\uparrow$ & LPIPS\,$\downarrow$ & tLPIPS\,$\downarrow$ & {} \\
\midrule
{} & {} & \FlexVSRctx{LiteVSR$^{N}$~\cite{pmlr-v306-cao26n} (ICML 2026)} & \underline{26.79} & 0.1943 & \textbf{0.01038} & \underline{25.97} & 0.2200 & \underline{0.01187} & \underline{2.83} \\
{} & \FlexVSRtype{-1.2ex}{Offline VSR\\methods} & \FlexVSRctx{LiteVSR$^{D}$ (ICML 2026)} & 23.06 & 0.3576 & 0.01470 & 23.35 & 0.3392 & 0.01434 & 9.67 \\
{} & {} & \FlexVSRctx{LUVE$^{N}$~\cite{pmlr-v306-zhao26d} (ICML 2026)} & 16.64 & 0.4398 & 0.02030 & 16.32 & 0.4540 & 0.02222 & 12.00 \\
{} & {} & \FlexVSRctx{VEnhancer$^{N}$~\cite{he2024venhancer} (arXiv 2024)} & 19.90 & 0.3274 & 0.01736 & 20.15 & 0.3254 & 0.01798 & 10.50 \\
\cline{2-10}
{} & {} & \FlexVSRctx{FlashVSR-T$^{N}$~\cite{Zhuang_2026_CVPR} (CVPR 2026)} & 26.51 & 0.1723 & 0.01440 & 25.72 & 0.1851 & 0.01421 & 4.33 \\
\raisebox{-1.2ex}[0pt][0pt]{\textbf{YouHQ40}} & {} & \FlexVSRctx{FlashVSR-T$^{D}$ (CVPR 2026)} & 23.65 & 0.2209 & 0.01426 & 23.67 & 0.2206 & 0.01419 & 6.67 \\
{} & {} & \FlexVSRctx{FlashVSR-F$^{N}$ (CVPR 2026)} & 26.04 & 0.1805 & 0.01456 & 25.26 & 0.1919 & 0.01374 & 4.67 \\
{} & \FlexVSRtype{-1.2ex}{Streaming VSR\\methods} & \FlexVSRctx{FlashVSR-F$^{D}$ (CVPR 2026)} & 23.17 & 0.2267 & \underline{0.01101} & 23.20 & 0.2264 & \textbf{0.01096} & 6.17 \\
{} & {} & \FlexVSRctx{InfVSR$^{N}$~\cite{pmlr-v306-zhang26dm} (ICML 2026)} & 24.47 & 0.2981 & 0.01896 & 24.72 & 0.2483 & 0.01581 & 8.67 \\
{} & {} & \FlexVSRctx{InfVSR$^{D}$ (ICML 2026)} & 24.72 & 0.2289 & 0.01400 & 24.92 & 0.2236 & 0.01435 & 6.67 \\
{} & {} & \FlexVSRctx{Ultra Flash$^{N}$~\cite{UltraFlash_2026} (NeurIPS 2026)} & 25.10 & \underline{0.1405} & 0.01298 & 24.14 & \underline{0.1699} & 0.01416 & 4.17 \\
\cline{3-10}
{} & {} & \FlexVSRctx{FlexVSR (Ours)} & \textbf{27.25} & \textbf{0.1122} & 0.01244 & \textbf{26.79} & \textbf{0.1277} & 0.01242 & \textbf{1.67} \\
\midrule
{} & {} & \FlexVSRctx{LiteVSR$^{N}$~\cite{pmlr-v306-cao26n} (ICML 2026)} & \underline{27.22} & 0.1453 & 0.00802 & 26.34 & 0.1802 & 0.00961 & 4.83 \\
{} & \FlexVSRtype{-1.2ex}{Offline VSR\\methods} & \FlexVSRctx{LiteVSR$^{D}$ (ICML 2026)} & 22.89 & 0.2896 & 0.01276 & 23.22 & 0.2739 & 0.01247 & 10.83 \\
{} & {} & \FlexVSRctx{LUVE$^{N}$~\cite{pmlr-v306-zhao26d} (ICML 2026)} & 18.88 & 0.3427 & 0.01145 & 18.74 & 0.3652 & 0.01273 & 11.83 \\
{} & {} & \FlexVSRctx{VEnhancer$^{N}$~\cite{he2024venhancer} (arXiv 2024)} & 20.03 & 0.2696 & 0.01011 & 20.33 & 0.2732 & 0.01060 & 10.00 \\
\cline{2-10}
{} & {} & \FlexVSRctx{FlashVSR-T$^{N}$~\cite{Zhuang_2026_CVPR} (CVPR 2026)} & \textbf{27.48} & \underline{0.1067} & 0.00845 & \textbf{26.43} & \underline{0.1253} & 0.00893 & \underline{2.83} \\
\raisebox{-1.2ex}[0pt][0pt]{\textbf{Inter4K-40}} & {} & \FlexVSRctx{FlashVSR-T$^{D}$ (CVPR 2026)} & 23.86 & 0.1810 & 0.01101 & 23.88 & 0.1809 & 0.01099 & 8.67 \\
{} & {} & \FlexVSRctx{FlashVSR-F$^{N}$ (CVPR 2026)} & 27.08 & 0.1107 & \underline{0.00745} & 26.06 & 0.1282 & \textbf{0.00772} & \underline{2.83} \\
{} & \FlexVSRtype{-1.2ex}{Streaming VSR\\methods} & \FlexVSRctx{FlashVSR-F$^{D}$ (CVPR 2026)} & 23.44 & 0.1813 & 0.00980 & 23.46 & 0.1812 & 0.00978 & 8.67 \\
{} & {} & \FlexVSRctx{InfVSR$^{N}$~\cite{pmlr-v306-zhang26dm} (ICML 2026)} & 26.14 & 0.1362 & 0.00838 & 26.10 & 0.1329 & 0.00808 & 4.33 \\
{} & {} & \FlexVSRctx{InfVSR$^{D}$ (ICML 2026)} & 25.05 & 0.1471 & 0.00938 & 25.26 & 0.1517 & 0.00935 & 6.50 \\
{} & {} & \FlexVSRctx{Ultra Flash$^{N}$~\cite{UltraFlash_2026} (NeurIPS 2026)} & 25.20 & 0.1160 & 0.00788 & 24.25 & 0.1477 & 0.00855 & 4.83 \\
\cline{3-10}
{} & {} & \FlexVSRctx{FlexVSR (Ours)} & 26.17 & \textbf{0.1040} & \textbf{0.00723} & \underline{26.38} & \textbf{0.1184} & \underline{0.00801} & \textbf{1.83} \\
\bottomrule
\end{tabular}
}
\ifdim\wd0>\linewidth
  \resizebox{\linewidth}{!}{\usebox0}%
\else
  \usebox0
\fi

\par\vspace{2pt}\begin{minipage}{\linewidth}
$N/D$ follow Figure~\ref{fig:teaser}; \text{-T}/\text{-F} denote Tiny/Full decoders for FlashVSR. Avg.\ Rank is averaged over all 6 columns per dataset.
\end{minipage}
\end{table}

Table~\ref{tab:quality_comparison} compares reconstruction and temporal
quality across datasets and scales. FlexVSR achieves the best LPIPS across all
four dataset--scale settings and the highest PSNR on YouHQ40 (+0.46/0.82~dB
over the strongest baseline LiteVSR$^{N}$ and +0.74/1.07~dB over
FlashVSR-T$^{N}$ at $1.5\times/2\times$). On Inter4K-40, FlexVSR ranks second
in $2\times$ PSNR (26.38~dB, surpassing LiteVSR$^{N}$ and trailing
FlashVSR-T$^{N}$ by only 0.05~dB). While native-$4\times$ restoration yields higher PSNR for FlashVSR-T$^\text{N}$ (+1.31/0.05\,dB at $1.5\times$/$2\times$) and for LiteVSR$^\text{N}$ and FlashVSR-F$^\text{N}$ at $1.5\times$ (+1.05 and +0.91\,dB), it is $1.5\text{--}37\times$ slower among streaming baselines (Table~\ref{tab:efficiency}, Sec.~4.3), while offline LiteVSR$^\text{N}$ takes 240/237\,s for a 13-frame prefix ($>120\times$ slower). For
temporal consistency (tLPIPS), FlexVSR ranks first/second at $1.5\times/2\times$
on Inter4K-40 and third at both scales on YouHQ40. Averaged across all six
metric columns, FlexVSR achieves the best overall rank on both YouHQ40 (1.67
vs.\ 2.83 for LiteVSR$^{N}$ and 4.17 for Ultra~Flash$^{N}$) and Inter4K-40
(1.83 vs.\ 2.83 for FlashVSR-T$^{N}$/-F$^{N}$) (Table~\ref{tab:quality_comparison}).

\noindent\textbf{Fixed-LR evaluation.} Complementing Table~\ref{tab:quality_comparison}'s fixed-HR setup, we upscale each fixed-res\-o\-lu\-tion YouHQ40 LR clip to $1.5\times$, unseen $1.8\times$, and $2.0\times$ target grids with the same weights (Appendix~\ref{sec:supp_fixed_lr}). FlexVSR achieves the lowest LPIPS at all scales ($0.1101/0.1406/0.1277$), best $1.8\times$ tLPIPS ($0.00877$) and $2.0\times$ PSNR ($26.79$~dB), and best average rank ($1.78$) over all nine scale-metric pairs.

\subsection{Efficiency and Adaptation Cost}
\label{sec:efficiency_analysis}
\begin{table}[!htb]
\centering
\setlength{\aboverulesep}{0pt}\setlength{\belowrulesep}{0pt}
\caption{\textbf{Streaming SR efficiency} from LR latents (w/ required decoding/resizing). Latency (in s) uses $t_{16}=(T_{77}-T_{13})/4$.}
\label{tab:efficiency}
\fontsize{9}{11.0}\selectfont
\setlength{\tabcolsep}{2.2pt}

\renewcommand{\arraystretch}{1.15}
\begin{tabular*}{\linewidth}{@{}l@{\extracolsep{\fill}}*{6}{c}@{}}
\toprule
\multicolumn{1}{@{}c}{\multirow{2}{*}{Method}} &
\multicolumn{3}{c}{$1.5\times$} & \multicolumn{3}{c@{}}{$2\times$} \\
\cmidrule(lr){2-4}\cmidrule(lr){5-7}
{} & $T_{13}\!\downarrow$ & $t_{16}\!\downarrow$ & FPS$\uparrow$
& $T_{13}\!\downarrow$ & $t_{16}\!\downarrow$ & FPS$\uparrow$ \\
\midrule
FlashVSR-T$^{N}$ & 10.335 & 6.408 & 2.50 & 5.678 & 3.588 & 4.46 \\
FlashVSR-T$^{D}$ & 2.353 & 1.870 & 8.55 & 1.890 & 1.411 & 11.34 \\
FlashVSR-F$^{N}$ & 18.994 & 14.105 & 1.13 & 9.564 & 6.188 & 2.59 \\
FlashVSR-F$^{D}$ & 3.402 & 2.563 & 6.24 & 2.948 & 2.138 & 7.48 \\
InfVSR$^{N}$ & 97.299 & 89.151 & 0.18 & 57.139 & 52.347 & 0.31 \\
InfVSR$^{D}$ & 2.375 & 2.815 & 5.68 & 1.892 & 2.376 & 6.73 \\
Ultra Flash$^{N}$ & \underline{0.885} & \underline{0.713} & \underline{22.44} & \underline{0.476} & \textbf{0.467} & \textbf{34.27} \\
\midrule
FlexVSR (Ours) & 1.957 & 2.387 & 6.70 & 1.885 & 2.388 & 6.70 \\
FlexVSR + TAEHV & \textbf{0.389} & \textbf{0.674} & \textbf{23.73} & \textbf{0.385} & \underline{0.668} & \underline{23.95} \\
\bottomrule
\end{tabular*}
\end{table}

Mirroring a coarse-to-fine generation pipeline where upstream synthesis directly outputs LR latents, Table~\ref{tab:efficiency} benchmarks $704\times1280$ streaming efficiency on one H100 GPU over a 77-frame YouHQ40 clip ($464\times848$/$352\times640$ LR at $1.5\times$/$2\times$), including LR decoding for RGB restorers while latent cascades decode once at HR. Native-$4\times$ routes incur heavy 16-frame latency that worsens at $1.5\times$ ($t_{16}=6.408/3.588$~s for FlashVSR-T$^{N}$, $14.105/6.188$~s for FlashVSR-F$^{N}$, and $89.151/52.347$~s for InfVSR$^{N}$), and FlashVSR-F$^{N}$ needs manual DiT cache cleanup before decoding to avoid $1.5\times$ out-of-memory errors. Though target-grid execution stabilizes cross-scale runtime, all D routes still take longer than default FlexVSR over the 13-frame prefix ($T_{13}$) and sharply degrade quality (Table~\ref{tab:quality_comparison}).
For adaptation cost, Stage-2 ($7.3$ H100 GPU-hours) avoids repeating Stage-1 prior training (${\sim}17$ H100 GPU-hours).

\noindent\textbf{Decoder choice.} Default Wan pixel decoding dominates runtime ($t_{16}=2.387/2.388$~s, $6.70/\allowbreak6.70$~FPS at $1.5\times$/$2\times$), whereas fast baselines use lightweight decoders like FlashVSR-T's Tiny decoder or TAEHV~\cite{BoerBohan2025TAEHV}. Keeping the frozen Wan~\cite{wan2025} latent space lets FlexVSR plug in TAEHV without retraining, giving the lowest prefix latency $T_{13}$ overall ($0.389/0.385$~s) and cutting $t_{16}$ to $0.674/0.668$~s ($23.73/23.95$~FPS, $3.54/3.57\times$ speedup; Fig.~\ref{fig:teaser}b for $1.5\times$). With a modest YouHQ40 trade-off ($26.38/25.98$~dB PSNR), it still beats all compared D-route baselines on all Inter4K-40 metrics. While fixed-scale optimizations give Ultra Flash higher native $2\times$ throughput ($34.27$ vs.\ $23.95$~FPS)~\cite{UltraFlash_2026}, target-grid execution without DiT acceleration lets FlexVSR + TAEHV surpass it at $1.5\times$ ($23.73$ vs.\ $22.44$~FPS) with $+1.28/1.84$~dB higher PSNR and lower $T_{13}$ at both scales. Further, our historical KV-cache variant (FlexVSR-KV) avoids recomputing history, cutting TAEHV $t_{16}$ to 0.562/0.558 s (28.44/28.69 FPS) without PSNR drop (TAEHV/KV-cache in Appendix~\ref{sec:supp_decoder}).

\subsection{Continuous-Scale Generalization}
\label{sec:scale_generalization}
\begin{table}[!htb]
\centering
\setlength{\aboverulesep}{0pt}\setlength{\belowrulesep}{0pt}
\caption{\textbf{Cross-scale comparison on YouHQ40} (supervised at $1.5{\times}/2{\times}$, matching Table~\ref{tab:quality_comparison}; Actual $H/W$ lists ratios for the 37 landscape clips). Best/second-best scores are \textbf{bold}/\underline{underlined}.}
\label{tab:cross_scale}
\fontsize{9}{11.0}\selectfont
\setlength{\tabcolsep}{2.5pt}

\renewcommand{\arraystretch}{1.18}
\newcommand{\FlexVSRmetric}[1]{\raisebox{-1.2ex}[0pt][0pt]{\itshape\begin{tabular}[c]{@{}c@{}}#1\end{tabular}}}
\sbox0{%
\begin{tabular}{c@{\hspace{3pt}}|@{\hspace{3pt}}l@{\extracolsep{0pt}}*{10}{c}@{}}
\toprule
\multirow{2}{*}{Metric} & Method & $1.2\times$ & $1.25\times$ & $1.3\times$ & $1.4\times$ & $1.5\times$ & $1.6\times$ & $1.7\times$ & $1.75\times$ & $1.8\times$ & $2\times$ \\
{} & Actual $H/W$ & 1.20/1.20 & 1.25/1.25 & 1.29/1.30 & 1.40/1.40 & 1.50/1.50 & 1.61/1.60 & 1.69/1.69 & 1.74/1.74 & 1.78/1.79 & 2.00/2.00 \\
\midrule
{} & FlashVSR-T$^{N}$\rule{0pt}{2.4ex} & \textbf{27.19} & \textbf{27.08} & \textbf{26.95} & \underline{26.77} & \underline{26.51} & \underline{26.33} & \underline{26.14} & \underline{26.08} & \underline{25.99} & \underline{25.72} \\
{} & FlashVSR-T$^{D}$ & 23.64 & 23.64 & 23.64 & 23.65 & 23.65 & 23.66 & 23.66 & 23.66 & 23.66 & 23.67 \\
\FlexVSRmetric{PSNR\,$\uparrow$} & InfVSR$^{N}$ & 24.21 & 24.24 & 24.29 & 24.42 & 24.47 & 24.53 & 24.57 & 24.59 & 24.62 & 24.72 \\
{} & InfVSR$^{D}$ & 24.53 & 24.56 & 24.60 & 24.66 & 24.72 & 24.78 & 24.82 & 24.83 & 24.85 & 24.92 \\
{} & Ultra Flash$^{N}$ & 25.80 & 25.65 & 25.54 & 25.29 & 25.10 & 24.85 & 24.71 & 24.60 & 24.52 & 24.14 \\
\cline{2-12}
{} & FlexVSR (Ours)\rule{0pt}{2.4ex} & \underline{26.42} & \underline{26.64} & \underline{26.83} & \textbf{27.03} & \textbf{27.25} & \textbf{26.83} & \textbf{26.54} & \textbf{26.37} & \textbf{26.20} & \textbf{26.79} \\
\midrule
{} & FlashVSR-T$^{N}$\rule{0pt}{2.4ex} & 0.1603 & 0.1629 & 0.1654 & 0.1677 & 0.1723 & 0.1747 & 0.1779 & 0.1790 & 0.1809 & 0.1851 \\
{} & FlashVSR-T$^{D}$ & 0.2209 & 0.2209 & 0.2208 & 0.2208 & 0.2209 & 0.2207 & 0.2207 & 0.2207 & 0.2207 & 0.2206 \\
\FlexVSRmetric{LPIPS\,$\downarrow$} & InfVSR$^{N}$ & 0.3212 & 0.3309 & 0.3289 & 0.3154 & 0.2981 & 0.2762 & 0.2645 & 0.2591 & 0.2549 & 0.2483 \\
{} & InfVSR$^{D}$ & 0.2384 & 0.2366 & 0.2350 & 0.2318 & 0.2289 & 0.2267 & 0.2255 & 0.2248 & 0.2242 & 0.2236 \\
{} & Ultra Flash$^{N}$ & \textbf{0.1225} & \textbf{0.1258} & \underline{0.1285} & \underline{0.1349} & \underline{0.1405} & \underline{0.1477} & \underline{0.1528} & \underline{0.1565} & \underline{0.1596} & \underline{0.1699} \\
\cline{2-12}
{} & FlexVSR (Ours)\rule{0pt}{2.4ex} & \underline{0.1366} & \underline{0.1292} & \textbf{0.1235} & \textbf{0.1178} & \textbf{0.1122} & \textbf{0.1224} & \textbf{0.1311} & \textbf{0.1371} & \textbf{0.1428} & \textbf{0.1277} \\
\midrule
{} & FlashVSR-T$^{N}$\rule{0pt}{2.4ex} & 0.01291 & 0.01351 & 0.01393 & 0.01409 & 0.01440 & 0.01436 & 0.01436 & 0.01428 & 0.01435 & 0.01421 \\
{} & FlashVSR-T$^{D}$ & 0.01430 & 0.01429 & 0.01428 & 0.01427 & 0.01426 & 0.01425 & 0.01425 & 0.01423 & 0.01423 & 0.01419 \\
\FlexVSRmetric{tLPIPS\,$\downarrow$} & InfVSR$^{N}$ & 0.02116 & 0.02149 & 0.02123 & 0.02029 & 0.01896 & 0.01767 & 0.01688 & 0.01653 & 0.01628 & 0.01581 \\
{} & InfVSR$^{D}$ & 0.01412 & 0.01408 & 0.01406 & 0.01401 & 0.01400 & 0.01403 & 0.01408 & 0.01411 & 0.01414 & 0.01435 \\
{} & Ultra Flash$^{N}$ & \underline{0.01199} & \underline{0.01220} & \underline{0.01234} & \underline{0.01267} & \underline{0.01298} & \underline{0.01328} & \underline{0.01341} & \underline{0.01358} & \underline{0.01371} & \underline{0.01416} \\
\cline{2-12}
{} & FlexVSR (Ours)\rule{0pt}{2.4ex} & \textbf{0.01151} & \textbf{0.01153} & \textbf{0.01142} & \textbf{0.01141} & \textbf{0.01244} & \textbf{0.01123} & \textbf{0.01090} & \textbf{0.01068} & \textbf{0.01048} & \textbf{0.01242} \\
\bottomrule
\end{tabular}
}
\ifdim\wd0>\linewidth
  \resizebox{\linewidth}{!}{\usebox0}%
\else
  \usebox0
\fi

\end{table}

\noindent\textbf{Cross-scale comparison.}
Table~\ref{tab:cross_scale} compares FlexVSR with streaming baselines (using
the stronger, faster FlashVSR-T) across ten YouHQ40 scales, covering
below-range extrapolation ($1.2\times$--$1.4\times$) and in-range interpolation
($1.6\times$--$1.8\times$). FlexVSR achieves the best tLPIPS at all ten scales,
best LPIPS across $1.3\times$--$2\times$ (second to Ultra~Flash$^{N}$ at
$1.2\times/1.25\times$), and highest PSNR across $1.4\times$--$2\times$ (second
to FlashVSR-T$^{N}$ at $1.2\times$--$1.3\times$). 
We further test above-range extrapolation ($2.1\times$--$2.2\times$) without
retraining, beyond both supervised scales and the frozen $2\times$ prior, where
FlexVSR still leads LPIPS and tLPIPS at both scales and ranks second in
$2.1\times$ PSNR (Appendix~\ref{sec:supp_extrapolation}).

\noindent\textbf{Wider-scale extension.}
With the $2\times$ prior frozen, a separate ReLSA adapter trained on 21 scales ($1.2\times$--$4\times$) evaluates low-latency scaling against the fastest streaming baselines, FlashVSR-T$^{D}$ and Ultra~Flash$^{N}$
(Table~\ref{tab:efficiency}, Fig.~\ref{fig:teaser}b). Across 17 YouHQ40 scales
(5 unseen), it achieves the best overall rank ($1.16$), topping all metrics in
$1.2\times$--$3\times$ and PSNR up to $3.5\times$ (see Appendix~\ref{sec:supp_wider_scales}). This broadens FlexVSR's practical applicability, allowing a single frozen $2\times$ prior to serve wide-ranging magnification demands without retraining the SR DiT.

\subsection{Ablation Studies}
\label{sec:component_ablation}
\begin{table}[!htb]
\centering
\setlength{\aboverulesep}{0pt}\setlength{\belowrulesep}{0pt}
\caption{\textbf{Ablation studies on YouHQ40} sharing the frozen Stage-1 Wan SR prior. Best and second-best scores are \textbf{bold} and \underline{underlined}.}
\label{tab:ablation_refinement}
\fontsize{9}{11.0}\selectfont
\setlength{\tabcolsep}{2pt}
\renewcommand{\arraystretch}{1.08}
\newcommand{\QRbox}[1]{{\fontsize{8}{10}\selectfont #1}}
\newcommand{\Mbox}[1]{\makebox[3.25em][c]{{\fontsize{8}{10}\selectfont #1}}} 
\sbox0{%
\begin{tabular}{@{}l@{\extracolsep{0pt}}*{3}{c@{\hspace{3pt}}}c@{\hspace{4pt}\extracolsep{0pt}}*{12}{c}@{}}
\toprule
\multicolumn{1}{@{}c}{\multirow{2}{*}{Variant}} & \multicolumn{4}{c}{Settings}
& \multicolumn{3}{c}{$1.2\times$ (unseen)}
& \multicolumn{3}{c}{$1.5\times$}
& \multicolumn{3}{c}{$1.8\times$ (unseen)}
& \multicolumn{3}{c@{}}{$2\times$} \\
\cmidrule(lr){2-5}\cmidrule(lr){6-8}\cmidrule(lr){9-11}\cmidrule(lr){12-14}\cmidrule(lr){15-17}
& \QRbox{Q} & \QRbox{R} & \QRbox{E2E} & \QRbox{SR}
& \Mbox{PSNR$\uparrow$} & \Mbox{LPIPS$\downarrow$} & \Mbox{DISTS$\downarrow$}
& \Mbox{PSNR$\uparrow$} & \Mbox{LPIPS$\downarrow$} & \Mbox{DISTS$\downarrow$}
& \Mbox{PSNR$\uparrow$} & \Mbox{LPIPS$\downarrow$} & \Mbox{DISTS$\downarrow$}
& \Mbox{PSNR$\uparrow$} & \Mbox{LPIPS$\downarrow$} & \Mbox{DISTS$\downarrow$} \\
\midrule
\multicolumn{17}{@{}l}{\textit{(a) Stage-2 adaptation data ablation (w/o VSR-17K for ReLSA; shared Stage-1 SR prior)}} \\
FlexVSR (Stage-2 REDS) & $\checkmark$ & $\checkmark$ & $\checkmark$ & $\checkmark$ & 25.99 & 0.1469 & 0.1016 & 26.98 & 0.1174 & 0.0821 & \underline{25.85} & 0.1539 & 0.1061 & 26.35 & 0.1448 & 0.1102 \\
\midrule
\multicolumn{17}{@{}l}{\textit{(b) Architecture ablations}} \\
w/o Query & $\times$ & $\checkmark$ & $\checkmark$ & $\checkmark$ & 25.37 & 0.1713 & 0.1108 & \underline{27.13} & \underline{0.1132} & \underline{0.0773} & 25.32 & 0.1518 & \underline{0.1004} & 26.58 & 0.1320 & 0.0967 \\
w/o Res & $\checkmark$ & $\times$ & $\checkmark$ & $\checkmark$ & \underline{26.40} & \underline{0.1413} & \textbf{0.0851} & 26.63 & 0.1266 & 0.0838 & 25.77 & \underline{0.1497} & 0.1046 & 25.93 & 0.1630 & 0.1164 \\
\midrule
\multicolumn{17}{@{}l}{\textit{(c) End-to-end (E2E) supervision and frozen SR prior (SR)}} \\
Warm-up (U-only) & $\checkmark$ & $\checkmark$ & $\times$ & $\times$ & 25.90 & 0.1684 & 0.1255 & 26.26 & 0.1553 & 0.1206 & 24.86 & 0.2091 & 0.1487 & 25.11 & 0.2324 & 0.1725 \\
E2E (U-only) & $\checkmark$ & $\checkmark$ & $\checkmark$ & $\times$ & 26.15 & 0.1439 & \underline{0.0858} & 26.19 & 0.1402 & 0.0892 & 24.51 & 0.1971 & 0.1330 & 25.02 & 0.1830 & 0.1344 \\
Warm-up + SR DiT & $\checkmark$ & $\checkmark$ & $\times$ & $\checkmark$ & 25.59 & 0.1728 & 0.1227 & 26.32 & 0.1351 & 0.1026 & 25.62 & 0.1700 & 0.1224 & 26.11 & 0.1632 & 0.1276 \\
\midrule
\multicolumn{17}{@{}l}{\textit{(d) Fixed-scale reference}} \\
Fixed $2\times$ U$^\dagger$ & -- & -- & -- & $\checkmark$ & 24.42 & 0.2358 & 0.1449 & 24.36 & 0.2191 & 0.1484 & 23.97 & 0.2131 & 0.1513 & \textbf{26.85} & \underline{0.1289} & \underline{0.0949} \\
\midrule
FlexVSR (Ours) & $\checkmark$ & $\checkmark$ & $\checkmark$ & $\checkmark$ & \textbf{26.42} & \textbf{0.1366} & 0.0921 & \textbf{27.25} & \textbf{0.1122} & \textbf{0.0755} & \textbf{26.20} & \textbf{0.1428} & \textbf{0.0944} & \underline{26.79} & \textbf{0.1277} & \textbf{0.0922} \\
\bottomrule
\end{tabular}
}
\ifdim\wd0>\linewidth
  \resizebox{\linewidth}{!}{\usebox0}%
\else
  \usebox0
\fi

\par\vspace{2pt}\begin{minipage}{\linewidth}\fontsize{9}{11.0}\selectfont
Q: continuous local query; R: learned target-grid refinement; E2E: end-to-end SR supervision via frozen DiT; SR: frozen SR DiT restoration at inference ($\times$: U-only, decoding $z_{\mathrm{up}}^{(s)}$ directly via the frozen VAE); U: latent upsampler. $\dagger$: frozen Stage-1 $2\times$ upsampler with bilinear latent resizing to target grids.
\end{minipage}
\end{table}

\noindent\textbf{Training data.}
With the Stage-1 prior fixed, adding VSR-17K over REDS-only in Stage-2
adaptation (Table~\ref{tab:ablation_refinement}(a)) improves PSNR by
$+0.27/+0.44$~dB at $1.5\times/2\times$ and $+0.43/+0.35$~dB at unseen
$1.2\times/1.8\times$, with consistently lower LPIPS and DISTS.

\noindent\textbf{Target-grid refinement.}
All variants in Table~\ref{tab:ablation_refinement}(b) follow the same
40k warm-up and 1k end-to-end training schedule, with identical
loss weights and data exposure.
Full ReLSA improves PSNR over removing the residual refinement branch (\emph{w/o Res}) by 0.62/0.86~dB at
$1.5\times/2\times$ and 0.02/0.43~dB at unseen $1.2\times/1.8\times$, with
consistently lower LPIPS across all four scales and lower DISTS at three scales
(\emph{w/o Res} is slightly lower only at $1.2\times$).
\noindent\textbf{Continuous querying.}
Replacing the bilinear base in \emph{w/o Query} with continuous local querying
improves PSNR by 0.12/0.21~dB at $1.5\times/2\times$ and by a substantial
1.05/0.88~dB at unseen $1.2\times/1.8\times$, accompanied by consistently
lower LPIPS and DISTS across all scales.

\noindent\textbf{End-to-end SR supervision and frozen prior.} 
In Table~\ref{tab:ablation_refinement}(c), adding 1k end-to-end (E2E) updates over the 40k warm-up (FlexVSR vs.\ Warm-up + SR DiT) boosts PSNR by $+0.93/+0.68$~dB at $1.5\times/2\times$ and $+0.83/+0.58$~dB at unseen $1.2\times/1.8\times$, with consistently lower LPIPS and DISTS. 
Evaluating upsampler-only (U-only) variants (decoding $z_{\mathrm{up}}^{(s)}$ directly via $D_\omega$ without the SR DiT) reveals that \textbf{standalone latent reconstruction does not ensure compatibility with a frozen generative prior}: at $1.2\times$, feeding unaligned warm-up latents into the frozen $2\times$ DiT even drops PSNR by $0.31$~dB vs.\ direct VAE decoding ($25.59$ vs.\ $25.90$~dB). Conversely, E2E optimization through the frozen restorer learns prior-compatible latent inputs, trading slight direct-decoding PSNR at $1.5\times\text{--}2\times$ for $+0.27\text{--}1.77$~dB SR gains while matching Fixed $2\times$ U at $2\times$ with superior perceptual quality (Table~\ref{tab:ablation_refinement}(d)).

\begin{table}[!htb]
\centering
\setlength{\aboverulesep}{0pt}\setlength{\belowrulesep}{0pt}
\caption{Cross-backbone and external-prior applicability on YouHQ40. Full-frame post-SR RGB quality against original HR RGB, averaged over 40 videos per scale (PSNR in dB). Upper block: priors prepared by us, each with its own VAE decoder. Last row: ReLSA on the frozen, independently released Ultra Flash prior, decoded with its TAEHV decoder.}
\label{tab:cross_dit}
\fontsize{9}{11.0}\selectfont\setlength{\tabcolsep}{3pt}\renewcommand{\arraystretch}{1.05}
\sbox0{%
\begin{tabular}{@{\extracolsep{0pt}}l*{12}{c}@{}}
\toprule
\multicolumn{1}{c}{\multirow{2}{*}{Prior / model}} & \multicolumn{3}{c}{$1.2\times$ (unseen)} & \multicolumn{3}{c}{$1.5\times$}
& \multicolumn{3}{c}{$1.8\times$ (unseen)} & \multicolumn{3}{c}{$2\times$} \\
\cmidrule(lr){2-4}\cmidrule(lr){5-7}\cmidrule(lr){8-10}\cmidrule(lr){11-13}
{} & PSNR$\uparrow$ & LPIPS$\downarrow$ & tLPIPS$\downarrow$ & PSNR$\uparrow$ & LPIPS$\downarrow$ & tLPIPS$\downarrow$
& PSNR$\uparrow$ & LPIPS$\downarrow$ & tLPIPS$\downarrow$ & PSNR$\uparrow$ & LPIPS$\downarrow$ & tLPIPS$\downarrow$ \\
\midrule
Wan2.1-1.3B (main) & 26.42 & 0.1366 & 0.01151 & 27.25 & 0.1122 & 0.01244 & 26.20 & 0.1428 & 0.01048 & 26.79 & 0.1277 & 0.01242 \\
Wan2.1-14B & 26.46 & 0.1326 & 0.01216 & 27.51 & 0.1072 & 0.01181 & 26.27 & 0.1387 & 0.01006 & 27.10 & 0.1219 & 0.01195 \\
MiniMax-H3 & 23.47 & 0.2001 & 0.01429 & 23.86 & 0.1883 & 0.01409 & 22.61 & 0.2231 & 0.01470 & 23.74 & 0.2086 & 0.01487 \\
\midrule
Ultra Flash + ReLSA & 25.69 & 0.1601 & 0.01260 & 26.19 & 0.1525 & 0.01347 & 26.03 & 0.1518 & 0.01353 & 26.00 & 0.1544 & 0.01392 \\
\bottomrule
\end{tabular}
}
\ifdim\wd0>\linewidth
  \resizebox{\linewidth}{!}{\usebox0}%
\else
  \usebox0
\fi

\end{table}
\noindent\textbf{Cross-backbone and external-prior applicability.}
Table~\ref{tab:cross_dit} evaluates separately trained ReLSA heads across frozen priors of varying scales, latent compression ratios, and external training pipelines. Under identical Stage-1/2 protocols, scaling to \textbf{Wan2.1-14B}~\cite{wan2025} improves PSNR by $0.04\text{--}0.31$~dB and lowers LPIPS across all scales. On \textbf{MiniMax-H3}~\cite{MiniMax_2026_H3} ($16\times$ spatial VAE vs.\ $8\times$ in Wan2.1), lower absolute PSNR primarily reflects its VAE reconstruction ceiling ($30.68$ vs.\ $32.66$~dB HR encode-decode on YouHQ40), while unseen $1.2\times/1.8\times$ scales stay within $1.25$~dB of supervised scales. To verify independence from our Stage-1 preparation, we also attach ReLSA directly to the externally released, frozen \textbf{Ultra Flash}~\cite{UltraFlash_2026} $2\times$ model by replacing only its final latent-upsampler output layer ($0.263$M parameters). Trained in just \textbf{4.7 H100 GPU-hours}, Ultra Flash + ReLSA surpasses the official Ultra Flash$^N$ route by \textbf{+1.09\text{--}1.86~dB} at $1.5\times\text{--}2\times$ (Table~\ref{tab:cross_scale}) while reaching \textbf{30.74/30.90~FPS} under Table~\ref{tab:efficiency}'s protocol, with E2E optimization through the frozen DiT contributing $+1.83\text{--}2.31$~dB over latent warm-up alone.

\subsection{Residual Analysis and Frozen-Prior Compatibility}
\label{sec:prior_compatibility}

\begingroup
\widowpenalty=10000
\paragraph{What the end-to-end phase contributes.}
The end-to-end phase differs from warm-up in two ways: gradients pass through the frozen SR DiT, and Eq.~(\ref{eq:supp_loss_e2e}) (Appendix~\ref{sec:supp_training}) adds pixel, perceptual, and temporal terms. To separate them, we resume from the warm-up checkpoint for 1,000 updates under identical loss weights, data order, windows, and noise: one variant backpropagates through the frozen SR DiT as in the main model, while the other bypasses it, evaluating every term on $z_{\mathrm{up}}^{(s)}$ and $\mathcal{D}(z_{\mathrm{up}}^{(s)})$ (Table~\ref{tab:e2e_control_main}, three-seed means). Adding decoded losses without the restorer recovers 0.36--0.48~dB of the 0.61--0.91~dB end-to-end PSNR gain, whereas keeping the frozen SR DiT in the loop adds a further $0.46/0.45/0.17/0.20$~dB at $1.2{\times}/1.5{\times}/1.8{\times}/2{\times}$ and lowers LPIPS at every scale (for all three seeds at each scale). Crucially, this ordering reverses without the restorer at inference: decoded directly, the bypass adapter is higher by 0.03--0.53~dB across scales and stays closer to $Y_s$. Through-restorer training thus trades standalone fidelity of $z_{\mathrm{up}}^{(s)}$ for downstream restoration quality.
\par\endgroup

\begin{table}[H]
\centering
\setlength{\aboverulesep}{0pt}\setlength{\belowrulesep}{0pt}
\caption{\textbf{Disentangling E2E loss terms vs.\ through-restorer gradients on YouHQ40} (40 videos per scale, full-frame RGB against original HR; PSNR in dB). Controlled rows report three-seed means (\textbf{bold}: better of the two) evaluating Eq.~(\ref{eq:supp_loss_e2e}) (Appendix~\ref{sec:supp_training}) weights on $z_{\mathrm{up}}^{(s)}$ and $\mathcal{D}(z_{\mathrm{up}}^{(s)})$ (bypassed) or on the frozen SR DiT outputs (through-restorer); the warm-up and main-model rows are single-run checkpoints. ``Decoded directly'' skips the SR DiT at inference.}
\label{tab:e2e_control_main}
\label{tab:supp_e2e_control}
\fontsize{9}{11.0}\selectfont\setlength{\tabcolsep}{3pt}\renewcommand{\arraystretch}{1.04}
\sbox0{%
\begin{tabular}{@{\extracolsep{0pt}}l*{4}{c}@{\hspace{6pt}}*{4}{c}@{\hspace{6pt}}*{4}{c}@{}}
\toprule
\multirow{2}{*}{Adapter} & \multicolumn{4}{c}{Restored by the SR DiT: PSNR$\uparrow$} & \multicolumn{4}{c}{Restored by the SR DiT: LPIPS$\downarrow$} & \multicolumn{4}{c}{Decoded directly: PSNR$\uparrow$} \\
\cmidrule(lr){2-5}\cmidrule(lr){6-9}\cmidrule(lr){10-13}
{} & $1.2{\times}$ & $1.5{\times}$ & $1.8{\times}$ & $2{\times}$ & $1.2{\times}$ & $1.5{\times}$ & $1.8{\times}$ & $2{\times}$ & $1.2{\times}$ & $1.5{\times}$ & $1.8{\times}$ & $2{\times}$ \\
\midrule
Warm-up (latent loss only) & 25.59 & 26.32 & 25.62 & 26.11 & 0.1728 & 0.1351 & 0.1700 & 0.1632 & 25.90 & 26.26 & 24.86 & 25.11 \\
+ E2E loss, SR DiT bypassed & 25.95 & 26.78 & 26.06 & 26.59 & 0.1578 & 0.1232 & 0.1528 & 0.1391 & \textbf{26.19} & \textbf{26.64} & \textbf{25.13} & \textbf{25.41} \\
+ E2E loss, through the SR DiT & \textbf{26.41} & \textbf{27.23} & \textbf{26.23} & \textbf{26.79} & \textbf{0.1403} & \textbf{0.1142} & \textbf{0.1462} & \textbf{0.1292} & 26.16 & 26.31 & 24.60 & 25.08 \\
\midrule
FlexVSR (Ours), main model & 26.42 & 27.25 & 26.20 & 26.79 & 0.1366 & 0.1122 & 0.1428 & 0.1277 & 26.15 & 26.19 & 24.51 & 25.02 \\
\bottomrule
\end{tabular}
}
\ifdim\wd0>\linewidth
  \resizebox{\linewidth}{!}{\usebox0}%
\else
  \usebox0
\fi

\end{table}

\paragraph{Transfer across restorers.}
To test whether this adaptation is tied to the specific restorer used during training, Table~\ref{tab:supp_restorer_swap} (Appendix~\ref{sec:supp_residual}) swaps trained adapters between the Wan2.1-1.3B and Wan2.1-14B pipelines (Table~\ref{tab:cross_dit}) without retraining, enabled by their shared Wan VAE. Under the 14B restorer, the 1.3B adapter gains 0.49--0.82~dB from its end-to-end phase, matching the 14B pipeline's native adapter (0.49--0.81~dB). Under the 1.3B restorer, the 14B adapter retains a 0.40--0.70~dB gain versus 0.58--0.93~dB for the matched 1.3B adapter, leaving at most 0.23~dB specific to one restorer. Restorer compatibility is thus largely shared across SR DiTs prepared with the same Stage-1 recipe rather than tied to a single set of weights.

\begin{figure}[H]
  \centering
  \includegraphics[width=0.82\linewidth]{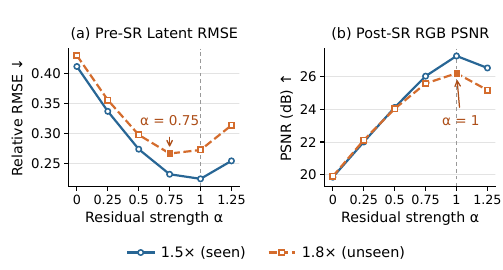}
  \caption{\textbf{Residual-strength intervention on YouHQ40} (40 videos) with fixed FlexVSR weights, varying $z_{\alpha}^{(s)} = z_{\mathrm{base}}^{(s)} + \alpha\Delta z^{(s)}$. At unseen $1.8\times$, pre-SR latent RMSE bottoms out at $\alpha=0.75$~(a) whereas post-SR PSNR peaks at $\alpha=1$~(b), confirming that $\Delta z^{(s)}$ calibrates inputs for frozen-DiT compatibility beyond mere latent proximity.}
  \label{fig:residual_scaling}
\end{figure}

\begin{figure}[H]
\centering
\includegraphics[width=0.98\textwidth]{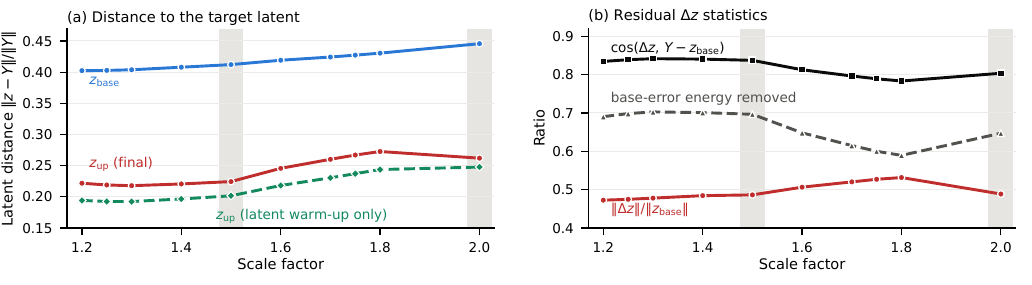}
\caption{\textbf{Residual analysis on YouHQ40} (40 videos, fixed weights; gray bands mark the supervised $1.5{\times}$ and $2{\times}$). (a) Relative $\ell_2$ distance to the encoded HR target $Y_s$ for the final model (solid) and its latent-only warm-up checkpoint of Table~\ref{tab:ablation_refinement}(c) (dashed). (b) Refiner residual statistics of the final model; base-error energy removed is $1-\|Y_s-z_{\mathrm{up}}^{(s)}\|_2^2/\|Y_s-z_{\mathrm{base}}^{(s)}\|_2^2$. Panels omit $s$ and $\ell_2$.}
\label{fig:supp_residual_stats}
\end{figure}

\paragraph{Residual strength and downstream restoration.}
Across $1.2\times$--$2\times$ on YouHQ40 (Fig.~\ref{fig:supp_residual_stats}), $\Delta z^{(s)}$ is non-trivial (RMS $0.47$--$0.53$ of $z_{\mathrm{base}}^{(s)}$), aligns closely with the base-to-target error $Y_s - z_{\mathrm{base}}^{(s)}$ (cosine similarity $0.78$--$0.84$), and removes $59$--$70\%$ of the squared base-error energy ($36$--$46\%$ lower relative $\ell_2$ distance to $Y_s$). Moreover, scaling the learned residual at test time ($z_{\alpha}^{(s)} = z_{\mathrm{base}}^{(s)} + \alpha\Delta z^{(s)}$, Fig.~\ref{fig:residual_scaling}) shows that $\alpha=0$ drops PSNR by $7.39/6.29$~dB at $1.5\times/1.8\times$ while $\alpha=1.25$ also degrades quality; notably, unseen $1.8\times$ pre-SR latent RMSE bottoms out at $\alpha = 0.75$ whereas post-SR PSNR peaks at $\alpha = 1$, confirming that the frozen restorer relies on a precisely calibrated residual. Meanwhile, E2E training deliberately trades closer target-latent $\ell_2$ proximity ($0.22$--$0.27$ vs.\ $0.19$--$0.25$ after warm-up; Fig.~\ref{fig:supp_residual_stats}a) for $+0.58$--$0.93$~dB higher post-SR PSNR (Table~\ref{tab:ablation_refinement}(c); see Appendix~\ref{sec:supp_residual} for further diagnostics).

\begin{figure}[H]
\centering
\includegraphics[width=\textwidth]{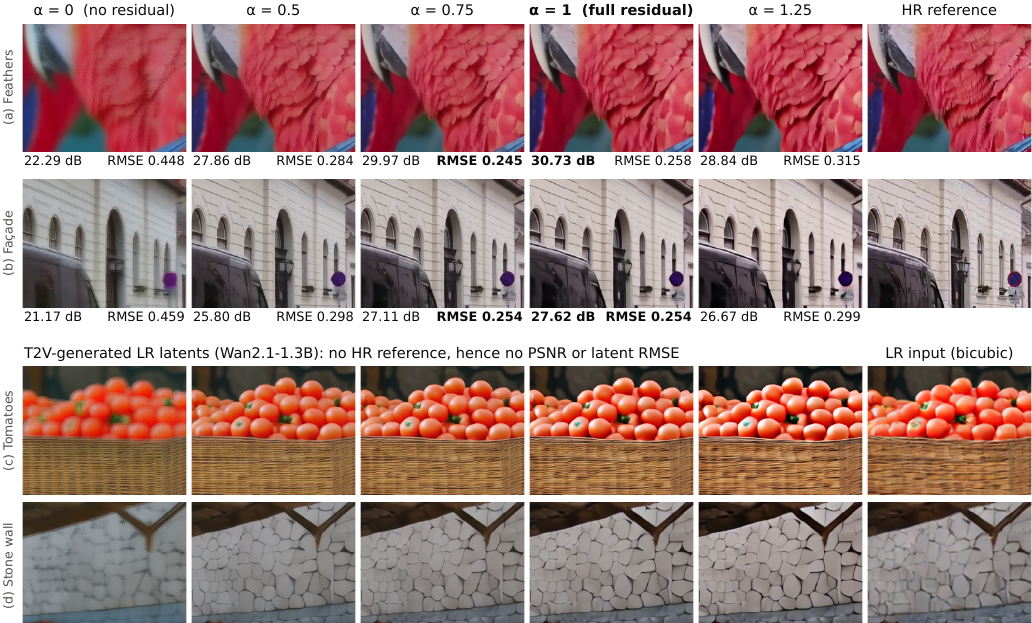}
\caption{\textbf{Residual-strength intervention at unseen $1.8{\times}$ on real videos and T2V-generated latents.} Varying $\alpha$ in $z_{\alpha}^{(s)}=z_{\mathrm{base}}^{(s)}+\alpha\Delta z^{(s)}$ ($256{\times}200$ native crops of frame 17; $\alpha=0.25$ omitted). (a, b) YouHQ40 bicubic LR runs (Fig.~\ref{fig:residual_scaling}) showing whole-clip post-SR PSNR and pre-SR latent RMSE (\textbf{bold}: best of six strengths; $\alpha=0.75, 1$ tie at three decimals in (b)). (c, d) Wan2.1-1.3B T2V latents ((d) from Fig.~\ref{fig:supp_qualitative_temporal}).}
\label{fig:supp_residual_cases}
\end{figure}

\paragraph{Per-video evidence and generated latents.}
Table~\ref{tab:supp_residual_pervideo} (Appendix~\ref{sec:supp_residual}) tests this contrast per video on the runs of Fig.~\ref{fig:residual_scaling} (fixed weights, input, and noise; only $\alpha$ changes). At supervised $1.5{\times}$, the two criteria largely agree: $\alpha=1$ gives lower latent RMSE than $\alpha=0.75$ on 36/40 videos and higher PSNR on all 40. At unseen $1.8{\times}$, they separate: latent RMSE is lower at $\alpha=0.75$ on 32 videos, yet PSNR is higher at $\alpha=1$ on 37 (the three exceptions differ by $\le$0.03~dB), and both hold on 29. The mismatch is thus not driven by a few samples. LPIPS also favors $\alpha=1$ on average (0.1428 vs.\ 0.1478; lower at $\alpha=1$ on 26/40 videos). On two T2V-generated LR latents without an HR reference (Fig.~\ref{fig:supp_residual_cases}c, d), visual comparison shows the same sensitivity: outputs are blurred without the residual, clean at $\alpha=1$, and show heavier contours at $\alpha=1.25$.

\FloatBarrier

\subsection{Qualitative Comparisons on Generated Latents}
\label{sec:qualitative_generated}

\begin{figure}[!htbp]
\centering
\includegraphics[width=\linewidth,height=\textheight,keepaspectratio]{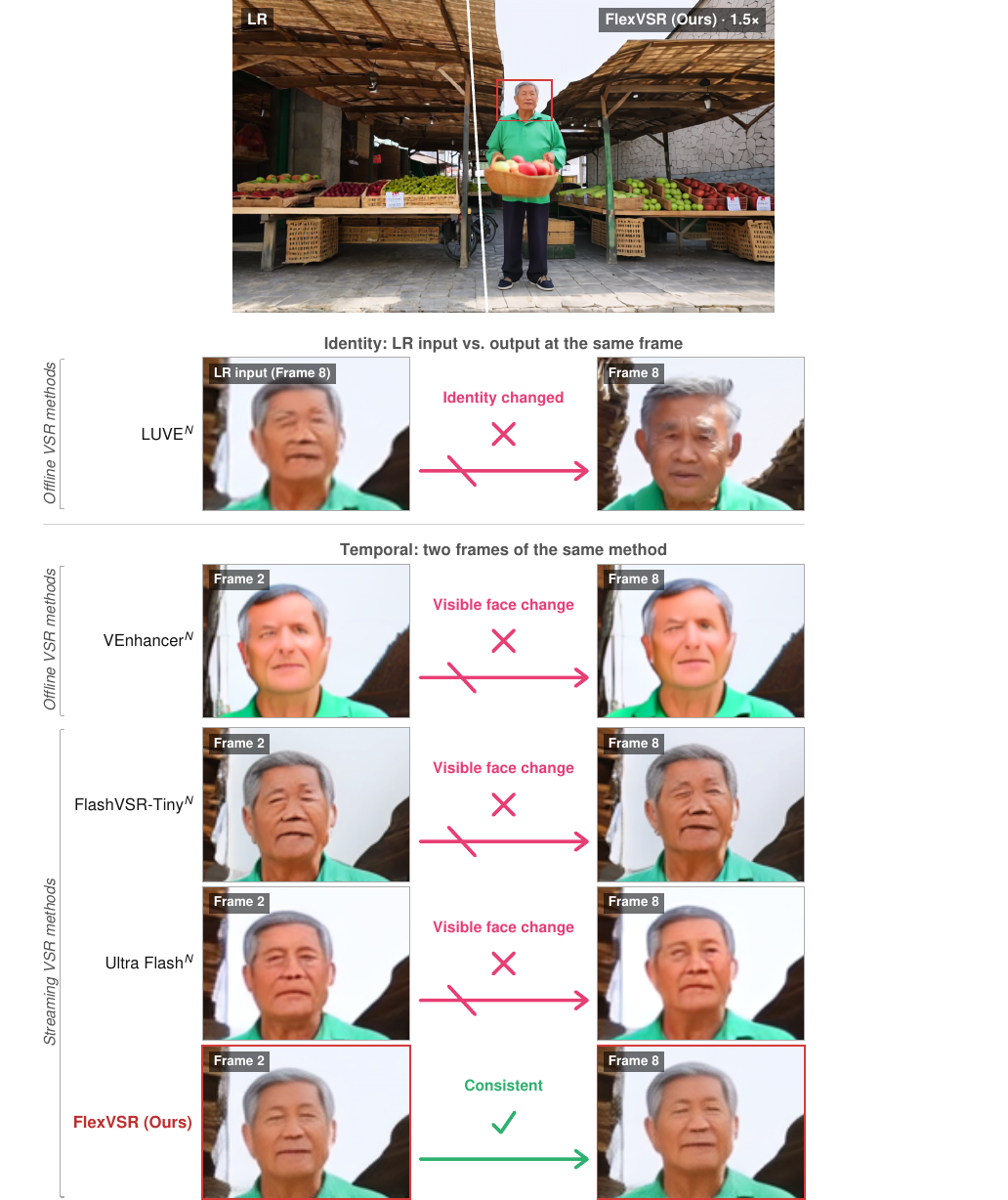}
\caption{\textbf{Temporal and identity consistency} at $1.5\times$ on a produce-stall video. All methods share the same upstream Wan2.1-1.3B-generated LR latent.}
\label{fig:supp_qualitative_temporal}
\end{figure}

Figure~\ref{fig:supp_qualitative_temporal} compares temporal and identity consistency at $1.5\times$ on a produce-stall video. All methods share the same upstream Wan2.1-1.3B-generated LR latent; continuous-scale comparisons on an additional scene are provided in Appendix~\ref{sec:supp_qualitative}.

\FloatBarrier

\FloatBarrier
\Needspace{15\baselineskip}
\section{Conclusion}
\label{sec:Conclusion}
\begingroup
\tolerance=999\emergencystretch=0pt
We present FlexVSR, unlocking practical continuous-scale capability for streaming generative VSR by adapting only the upstream latent interface of a frozen SR DiT.
Specifically, our $0.274$M-parameter ReLSA adapter is optimized end-to-end for frozen-prior compatibility in just $7.3$ H100 GPU-hours while preserving the original causal streaming protocol. 
Delivering favorable quality-latency trade-offs across unseen continuous scales, FlexVSR establishes a restoration paradigm that is flexible in both inference-time scaling and lightweight prior adaptation.
In practice, FlexVSR accommodates two complementary use cases: adapting varying low-resolution inputs to a fixed display resolution and mapping a fixed source stream to different target resolutions. 
This allows flexible scale adjustments offline or across streams without scale-specific retraining, while pointing toward dynamic scale switching during live generation and playback. We hope our work inspires future research on achieving both low latency and high visual quality in streaming and interactive video generation.
\par\endgroup

\par
{\small
  \setlength{\bibsep}{3pt}
  \bibliographystyle{ieeenat_fullname}
  \bibliography{main}
}
\clearpage
\appendix
\section*{Appendix}
\phantomsection
\addcontentsline{toc}{section}{Appendix}
\setcounter{table}{0}
\setcounter{figure}{0}
\setcounter{equation}{0}
\renewcommand{\thetable}{S\arabic{table}}
\renewcommand{\thefigure}{S\arabic{figure}}
\renewcommand{\theequation}{S\arabic{equation}}
\renewcommand{\theHtable}{supp.\arabic{table}}
\renewcommand{\theHfigure}{supp.\arabic{figure}}
\renewcommand{\theHequation}{supp.\arabic{equation}}
\noindent This appendix provides extended diagnostics, implementation and dataset specifications, and additional evaluations complementing the main paper:
\par\noindent \textbf{Sec.~\ref{sec:supp_residual}:} 
\textbf{Residual analysis on latent-error vs.\ post-SR quality mismatch across clips};
\par\noindent \textbf{Sec.~\ref{sec:supp_training}:} \textbf{Training schedules and loss weights}, including causal history sampling;
\par\noindent \textbf{Sec.~\ref{sec:supp_vsr17k}:} \textbf{VSR-17K curation and external-test overlap screening};
\par\noindent \textbf{Sec.~\ref{sec:supp_qualitative}:} \textbf{Additional continuous-scale qualitative comparisons} on generated latents;
\par\noindent \textbf{Sec.~\ref{sec:supp_fixed_lr}:} \textbf{Complementary fixed-LR evaluations} at multiple target resolutions;
\par\noindent \textbf{Sec.~\ref{sec:supp_decoder}:} \textbf{Full quality metrics for FlexVSR + TAEHV}, the timing protocol, and a faster KV-cache variant;
\par\noindent \textbf{Sec.~\ref{sec:supp_wider_scales}:} \textbf{Wider-scale experiments from $1.2\times$ to $4\times$};
\par\noindent \textbf{Sec.~\ref{sec:supp_routes}:} \textbf{Native/direct baseline routing};
\par\noindent \textbf{Sec.~\ref{sec:supp_full_metrics}:} \textbf{Extended metric comparison} of all main-table methods;
\par\noindent \textbf{Sec.~\ref{sec:supp_extrapolation}:} \textbf{Above-range extrapolation} without retraining.
\par\medskip

\FloatBarrier
\section{Residual Analysis: Latent Proximity vs.\ Frozen-Prior Compatibility}
\label{sec:supp_residual}
We analyze the final main model and its latent-only warm-up checkpoint from the same training run, both evaluated with fixed weights on YouHQ40 (40 videos, the ten scales of main-paper Table~\ref{tab:cross_scale}, $1.2{\times}$--$2{\times}$). We split the ReLSA output into base latent and refiner residual, $z_{\mathrm{up}}^{(s)}=z_{\mathrm{base}}^{(s)}+\Delta z^{(s)}$ (main-paper Eq.~(\ref{eq:relsa_refinement})), against the Wan-encoded HR target latent $Y_s$ (Sec.~\ref{sec:supp_training}; main-paper Sec.~\ref{sec:Decoder}), averaging full-resolution statistics over latent frames and videos.

\paragraph{Residual strength and downstream restoration.}
Fig.~\ref{fig:supp_residual_stats} plots the cross-scale latent distances (a) and refiner residual statistics (b) summarized in Sec.~\ref{sec:prior_compatibility}.
As a post-hoc diagnostic requiring the HR target (unused at inference), we compute per video and scale the residual gain minimizing pre-SR latent distance, ${\alpha^{\star}_{\mathrm{lat}}=\langle Y_s-z_{\mathrm{base}}^{(s)},\Delta z^{(s)}\rangle/\|\Delta z^{(s)}\|_2^2}$. Its mean over videos ranges from 0.83 to 0.97 across the ten scales, below full residual strength ($\alpha=1$). Among the strengths tested in main-paper Fig.~\ref{fig:residual_scaling}, pre-SR latent RMSE at $1.8{\times}$ is lowest at $\alpha=0.75$, whereas post-SR PSNR peaks at $\alpha=1$.

\paragraph{End-to-end training trades latent proximity for restorer compatibility.}
The latent-only warm-up checkpoint of main-paper Table~\ref{tab:ablation_refinement}(c) lies closer to $Y_s$ than the final model at every scale (relative $\ell_2$ distance 0.19--0.25 vs.\ 0.22--0.27; closer on $\ge$38/40 videos per scale; Fig.~\ref{fig:supp_residual_stats}a), yet achieves 0.58--0.93~dB lower PSNR through the same frozen SR DiT across $1.2{\times}/1.5{\times}/1.8{\times}/2{\times}$ (main-paper Table~\ref{tab:ablation_refinement}(c)). Together with the residual-strength results, this suggests that the learned latent interface benefits the frozen restorer beyond minimizing distance to $Y_s$.

\paragraph{Per-video statistics.}
Complementing Sec.~\ref{sec:prior_compatibility}, Table~\ref{tab:supp_residual_pervideo} details clip-level preference counts and PSNR gain quartiles for the $\alpha$-intervention on YouHQ40.

\begin{table}[!htbp]
\centering
\setlength{\aboverulesep}{0pt}\setlength{\belowrulesep}{0pt}
\caption{\textbf{Per-video statistics of the residual-strength intervention on YouHQ40} (40 videos per scale; main-paper Fig.~\ref{fig:residual_scaling} runs). ``Both'': lower latent RMSE at $\alpha=0.75$ and higher PSNR at $\alpha=1$.}
\label{tab:supp_residual_pervideo}
\fontsize{9}{11.0}\selectfont\setlength{\tabcolsep}{3pt}\renewcommand{\arraystretch}{1.04}
\sbox0{%
\begin{tabular}{@{\extracolsep{0pt}}l*{9}{c}@{}}
\toprule
\multirow{2}{*}{Scale} & \multicolumn{3}{c}{PSNR at $\alpha=1$ higher than at} & \multicolumn{2}{c}{\begin{tabular}[c]{@{}c@{}}Latent RMSE lower\\at $\alpha=0.75$\end{tabular}}
& \multicolumn{4}{c}{\begin{tabular}[c]{@{}c@{}}PSNR gain from $\alpha=0.75$\\to $\alpha=1$ (dB)\end{tabular}} \\
\cmidrule(lr){2-4}\cmidrule(lr){5-6}\cmidrule(lr){7-10}
{} & $\alpha=0$ & $\alpha=0.75$ & $\alpha=1.25$ & than at $\alpha=1$ & Both & mean & median & [$Q_1$, $Q_3$] & [min, max] \\
\midrule
$1.5{\times}$ (supervised) & 40 & 40 & 39 & 4 & 4 & $+1.23$ & $+1.06$ & [$+0.72$, $+1.50$] & [$+0.34$, $+3.18$] \\
$1.8{\times}$ (unseen) & 40 & 37 & 40 & 32 & 29 & $+0.62$ & $+0.46$ & [$+0.31$, $+0.77$] & [$-0.03$, $+1.88$] \\
\bottomrule
\end{tabular}
}
\ifdim\wd0>\linewidth
  \resizebox{\linewidth}{!}{\usebox0}%
\else
  \usebox0
\fi

\end{table}

\paragraph{Cross-restorer results.}
Table~\ref{tab:supp_restorer_swap} reports PSNR before and after E2E training for each adapter under both frozen restorers, complementing the transfer analysis in Sec.~\ref{sec:prior_compatibility}. The complete E2E loss control, including the single-run main-model checkpoint, is reported in Table~\ref{tab:e2e_control_main}.

\begin{table}[!htbp]
\centering
\setlength{\aboverulesep}{0pt}\setlength{\belowrulesep}{0pt}
\caption{\textbf{Restorer swap on YouHQ40} (40 videos per scale; PSNR in dB). Each adapter of the Wan2.1-1.3B and Wan2.1-14B pipelines (Table~\ref{tab:cross_dit}), before and after its end-to-end phase, is restored by either frozen SR DiT without retraining via their shared Wan VAE.}
\label{tab:supp_restorer_swap}
\fontsize{9}{11.0}\selectfont\setlength{\tabcolsep}{3pt}\renewcommand{\arraystretch}{1.04}
\sbox0{%
\begin{tabular}{@{\extracolsep{0pt}}l*{4}{c}@{\hspace{6pt}}*{4}{c}@{}}
\toprule
\multirow{2}{*}{Adapter} & \multicolumn{4}{c}{Restored by the 1.3B SR DiT} & \multicolumn{4}{c}{Restored by the 14B SR DiT} \\
\cmidrule(lr){2-5}\cmidrule(lr){6-9}
{} & $1.2{\times}$ & $1.5{\times}$ & $1.8{\times}$ & $2{\times}$ & $1.2{\times}$ & $1.5{\times}$ & $1.8{\times}$ & $2{\times}$ \\
\midrule
1.3B pipeline, warm-up & 25.59 & 26.32 & 25.62 & 26.11 & 25.75 & 26.58 & 25.76 & 26.45 \\
1.3B pipeline, end-to-end through 1.3B & 26.42 & 27.25 & 26.20 & 26.79 & 26.46 & 27.40 & 26.25 & 27.01 \\
\midrule
14B pipeline, warm-up & 25.55 & 26.45 & 25.66 & 26.15 & 25.74 & 26.70 & 25.78 & 26.47 \\
14B pipeline, end-to-end through 14B & 26.24 & 27.15 & 26.06 & 26.65 & 26.46 & 27.51 & 26.27 & 27.10 \\
\bottomrule
\end{tabular}
}
\ifdim\wd0>\linewidth
  \resizebox{\linewidth}{!}{\usebox0}%
\else
  \usebox0
\fi

\end{table}

\FloatBarrier
\section{Core Training Procedure}
\label{sec:supp_training}
The following recipe describes the main Wan2.1-1.3B~\cite{wan2025} model with sparse
scale supervision. Sec.~\ref{sec:supp_wider_scales} gives the sampling
scheme for the separate wider-scale ReLSA adapter. Notation follows the main paper.

\paragraph{Data, windows, and notation.}
We take 21-frame, $512\times512$ RGB crops and construct $256\times256$
LR inputs and $384\times384$/$512\times512$ targets using antialiased
bicubic resizing; Stage~1 uses only the $512\times512$ ($2\times$) targets.
Each resolution is encoded independently by the frozen Wan VAE, giving $Z_L$
and $Y_s$ with six latent frames. A training window restores four new latent
frames. With probability 0.25 it has no history, as for the first streaming
chunk; otherwise two history frames precede them:
\begin{equation}
\mathcal H=
\begin{cases}
\operatorname{sg}\big(\widehat Y^{\varnothing}_{1:2}\big) & \text{with probability } 0.75,\\
Y_{s,1:2} & \text{otherwise},
\end{cases}
\label{eq:supp_history}
\end{equation}
where $\widehat Y^{\varnothing}$ comes from a separate gradient-free forward pass
over the same six latent frames: the one-step prediction at $\sigma=0.2$ from
fresh noise and without history, by the SR DiT being trained (Stage~1) or by
the frozen SR DiT on the detached ReLSA output (Stage~2).
$\widehat Y^{\varnothing}_{1:2}$ are its outputs at latent positions 1--2 (RGB
frames 1--5), and $\operatorname{sg}$ stops gradients. The training pass then
restores positions 3--6 (RGB frames 6--21), or positions 1--4 when there is no
history; the history replaces the first two frames of $x_\sigma$ and receives a
clean timestep. Mixing ground-truth history with
detached self-predictions follows the principle of scheduled
sampling~\cite{Bengio_2015_NeurIPS} using a fixed mixing probability.
Losses on predictions are computed on the new frames with the Charbonnier
distance~\cite{Charbonnier_1997_TIP}
$\mathrm{Ch}(a,b)=\operatorname{mean}\sqrt{(a-b)^2+10^{-6}}$.
$\Delta_t$ denotes differences between adjacent frames. $\mathcal D(\cdot)$ is
short-crop decoding for the RGB losses: two adjacent latent frames of a $32\times32$ latent crop are decoded by the
frozen Wan VAE into four $224\times224$ RGB frames after border removal, and
$X=\mathcal D(Y_s)$.
LPIPS (AlexNet)~\cite{Zhang_2018_CVPR} and DISTS~\cite{ding2022dists} use
alternating decoded frames. Temporal-difference terms match frame-to-frame
changes to those of the target rather than penalizing them, which discourages
flicker without suppressing motion.

\paragraph{Stage~1: Reference-scale prior preparation.}
On VSR-17K, we first train the fixed-$2\times$ latent upsampler $U$
($F_\phi$ and the $2\times$ head) for 10,000 updates (global batch 32,
learning rate $2\!\times\!10^{-4}$). With the upsampler frozen, we then
adapt the SR DiT for 4,346 updates (global batch 8) using rank-16
LoRA~\cite{Hu_2022_LoRA} and trainable conditioning/output projections, with
learning rates $3\!\times\!10^{-4}$, $10^{-4}$, and $10^{-5}$ for LoRA, the LR
projection, and the output projection (cosine decay for LoRA and the output
projection). Each window builds $x_\sigma$ and $c_L$ by main-paper Eq.~(\ref{eq:sr_state_condition})
from $U(\tilde Z_L)$, where $\tilde Z_L$ is the perturbed LR latent below and
$\sigma\sim\mathcal U[0.1,0.3]$, and predicts
$\widehat Y_2=x_\sigma+\sigma\,v_\psi(x_\sigma,\sigma,c_L,p,\mathcal H)$ with the
velocity target $v^*=(Y_2-x_\sigma)/\sigma$ (main-paper Eq.~(\ref{eq:fm_prelim})).
The VAE remains frozen throughout.
Stage~1 is trained without dedicated
training acceleration and takes 17.4 H100 GPU-hours on eight GPUs: 10.35
for the upsampler and 7.04 for the SR DiT. As with the 7.3 H100
GPU-hours of Stage~2 in the main paper, we sum the training-step
time and multiply by the number of GPUs. This cost is paid once for the
reference-scale prior and is not part of the Stage-2 adaptation cost.

\paragraph{LR perturbation.}
During SR DiT adaptation, the LR latent $Z_L$ fed to the frozen upsampler
is perturbed (main-paper Fig.~2) as
$\tilde{Z}_L=Z_L-\beta\,m\,e\,(Z_L-\mathcal{B}Z_L)+\lambda\,m\,e\,s_z\,n+0.004\,s_z\,n'$,
with $s_z$ the per-channel standard deviation of training LR latents,
$\mathcal{B}$ a $5\times5$ spatial box filter, $m$ a soft elliptical mask
(all ones w.p.\ 0.3), $e\in(0,1]$ a temporal envelope, and $n,n'$ unit-RMS
noise mixing static, AR(1) ($\rho\sim\mathcal{U}[0.85,0.97]$), and i.i.d.\
fields with weights $(0.1,0.7,0.2)$.
Four levels, drawn with probabilities $(0.10,0.15,0.35,0.40)$, set
$\lambda\sim\mathcal{U}$ over $[0.005,0.015]$, $[0.015,0.04]$, $[0.04,0.08]$,
$[0.08,0.15]$; with probability $0.3/0.5/0.7/0.9$,
$\beta\sim\mathcal{U}$ over $[0,0.1]$, $[0.1,0.3]$, $[0.2,0.5]$, $[0.5,0.8]$
(otherwise $\beta=0$); with probability $0.3/0.5/0.6/0.7$, a per-channel
gain/bias acts on 30--70\% of channels; and $\tilde{Z}_L-Z_L$ is rescaled so
that its RMS is at most $0.04/0.10/0.18/0.28$ over the clip and
$0.07/0.16/0.28/0.45$ per frame, in units of $s_z$.

\paragraph{Stage~2: Latent-scale adaptation.}
Using REDS~\cite{nah2019ntire} and VSR-17K, we train only ReLSA, jointly for
both branches from random initialization (with the zero-initialized refiner
output of main-paper Eq.~(\ref{eq:relsa_refinement})), and supervise $s\in\{1.5,2\}$ equally.
A latent warm-up minimizes $\mathcal L_{\mathrm{warm}}=\mathrm{Ch}(z_{\mathrm{up}}^{(s)},Y_s)$
for 40,000 updates (global batch 32, learning rate $10^{-3}$). It is followed
by 1,000 end-to-end (E2E) updates through the frozen SR DiT (global batch 16,
learning rate $2\!\times\!10^{-4}$) at $\sigma=0.2$, minimizing
\begin{equation}
\begin{aligned}
\mathcal L_{\mathrm{E2E}}={}&\mathrm{Ch}(\widehat Y_s,Y_s)+5\,\mathrm{Ch}(\hat X,X)
+\mathrm{LPIPS}(\hat X,X)+0.5\,\mathrm{DISTS}(\hat X,X)\\
&+5\,\mathrm{Ch}(\Delta_t\hat X,\Delta_tX)
+0.05\,\mathrm{Ch}(\Delta_t\widehat Y_s,\Delta_tY_s)
+0.1\,\mathrm{Ch}(z_{\mathrm{up}}^{(s)},Y_s),
\end{aligned}
\label{eq:supp_loss_e2e}
\end{equation}
where $\widehat Y_s$ is the frozen one-step update of main-paper Sec.~\ref{sec:Decoder},
$\hat X=\mathcal D(\widehat Y_s)$, and the last term anchors the ReLSA output
$z_{\mathrm{up}}^{(s)}$ to the target latent. Gradients reach ReLSA through both $x_\sigma$ and
$c_L$ (main-paper Eq.~(\ref{eq:sr_state_condition})). The feature extractor, SR DiT, and Wan VAE remain
frozen. We use AdamW with weight decay $10^{-4}$, reset its state between
phases, warm up the learning rate for 100/50 updates, respectively, and clip
the gradient norm to 1.

\FloatBarrier
\section{VSR-17K for Generative Video Super-Resolution}
\label{sec:supp_vsr17k}

\paragraph{Motivation.}
Public generative VSR resources include diverse 1080p YouHQ
videos~\cite{Zhou_2024_CVPR} and quality- and motion-filtered
HQ-VSR~\cite{Chen_2025_DOVE}.
VSR-17K complements these resources with bitrate-screened 4K footage
and consistent target construction across scales.

\paragraph{Source and intended use.}
VSR-17K is curated from publicly available online videos and is intended
exclusively for non-commercial research. We will release the training
and validation pairs, split manifests, and paired-data construction
scripts to support reproducibility.

\paragraph{Curation.}
Source clips have recorded video-stream bitrates $\geq10$~Mbps and long
edges $\geq3{,}000$ pixels, predominantly $3840\times2160$
(Table~\ref{tab:vsr17k_statistics}); bitrate is a reproducible quality proxy.

\paragraph{Splits and paired supervision.}
We retain existing scene/temporal boundaries and timestamps, remove
duplicate clip IDs across pools, and split by source-video ID.
The final dataset contains 17,377 training clips from 3,493 source videos
and an additional 232 validation clips from 51 source-disjoint videos.
Both splits meet the bitrate criterion and share no source IDs or clip
IDs; deduplication is identifier-based.
For the main model, each pair contains 21 consecutive frames at the source
frame rate. LR and targets share the same time interval and spatial
region; each target is resized in RGB and independently VAE-encoded
(Sec.~\ref{sec:supp_training}). All 17,377 training clips are used for
Stage-1 prior preparation and, together with REDS, Stage-2 ReLSA adaptation.

\paragraph{Inter4K-40 test subset.}
Inter4K-40 is drawn from all 1,000 Inter4K videos~\cite{stergiou2023adapool}
and has the same number of clips as YouHQ40.
We sort the file names numerically, shuffle them with a fixed seed
(Python \texttt{random.Random(20261028)}), and keep the first 40 clips
whose first 90 frames contain no black frame; one clip (130) is skipped.
A frame counts as black if its mean or standard deviation is at most 4
on the 8-bit scale, measured on a $270\times480$ proxy.
The checked frames include all evaluated frames, and no other criterion
is applied.
Each clip is resized from $2160\times3840$ to $720\times1280$,
center-cropped to $704\times1280$ (height $\times$ width), and truncated
to its first 81 frames. LR inputs are bicubic-downsampled with
antialiasing to $464\times848$ and $352\times640$ for $1.5\times$ and
$2\times$.
The selected clip IDs are 18, 27, 77, 90, 115, 120, 169, 170, 173, 175, 203, 205, 210, 214, 263, 264, 275, 330, 342, 368, 369, 396, 407, 519, 530, 546, 575, 582, 617, 659, 696, 714, 763, 777, 836, 878, 900, 928, 944, 962.
We will release the selection script and this manifest.

\paragraph{External-test overlap screening.}
To check for content overlap against external benchmarks whose manifests
lack shared original-video identifiers, we performed a perceptual-hash
screen against all 40 YouHQ40 and 40 Inter4K-40 videos. Frames 0, 10,
and 20 from each training/validation pair's 21-frame source window,
before cropping (52,827 frames), were compared with each test video's
first, middle, and last HR frame (240 frames), using 64-bit DCT perceptual
hashes and Hamming threshold 8. No pair met the threshold (minimum
distance 10); inspecting the nearest pair per test video also found no
matching content. All 720 same-image checks under JPEG recompression,
half-resolution resizing, or brightness changes remained within the
threshold. Sampled-frame manifests, hashes, and nearest-pair records are
retained.

\begin{table}[!htbp]
\centering
\setlength{\aboverulesep}{0pt}\setlength{\belowrulesep}{0pt}
\caption{\textbf{VSR-17K statistics.} Duration/FPS describe source clips before
short-window extraction. $Q_1$/$Q_3$: quartiles.}
\label{tab:vsr17k_statistics}
\small\setlength{\tabcolsep}{3pt}\renewcommand{\arraystretch}{1.04}
\begin{tabular*}{\linewidth}{@{\extracolsep{\fill}}lcc@{}}
\toprule
Statistic & Training & Validation \\
\midrule
Clips / source videos & 17,377 / 3,493 & 232 / 51 \\
Video bitrate: median (Mbps) & 15.76 & 17.37 \\
Clip duration: median [$Q_1$, $Q_3$] (s) & 15.57 [9.01, 22.99] & 9.01 [9.00, 15.01] \\
Frame rate: median [min, max] (FPS) & 29.97 [20.42, 60] & 29.97 [23.98, 60] \\
$3840\times2160$ source clips & 16,792 (96.6\%) & 226 (97.4\%) \\
Frames per prepared pair (main model) & 21 & 21 \\
\bottomrule
\end{tabular*}
\end{table}

\clearpage
\section{Qualitative Comparisons}
\label{sec:supp_qualitative}
\begin{figure}[H]
\centering
\includegraphics[width=\linewidth,height=\textheight,keepaspectratio]{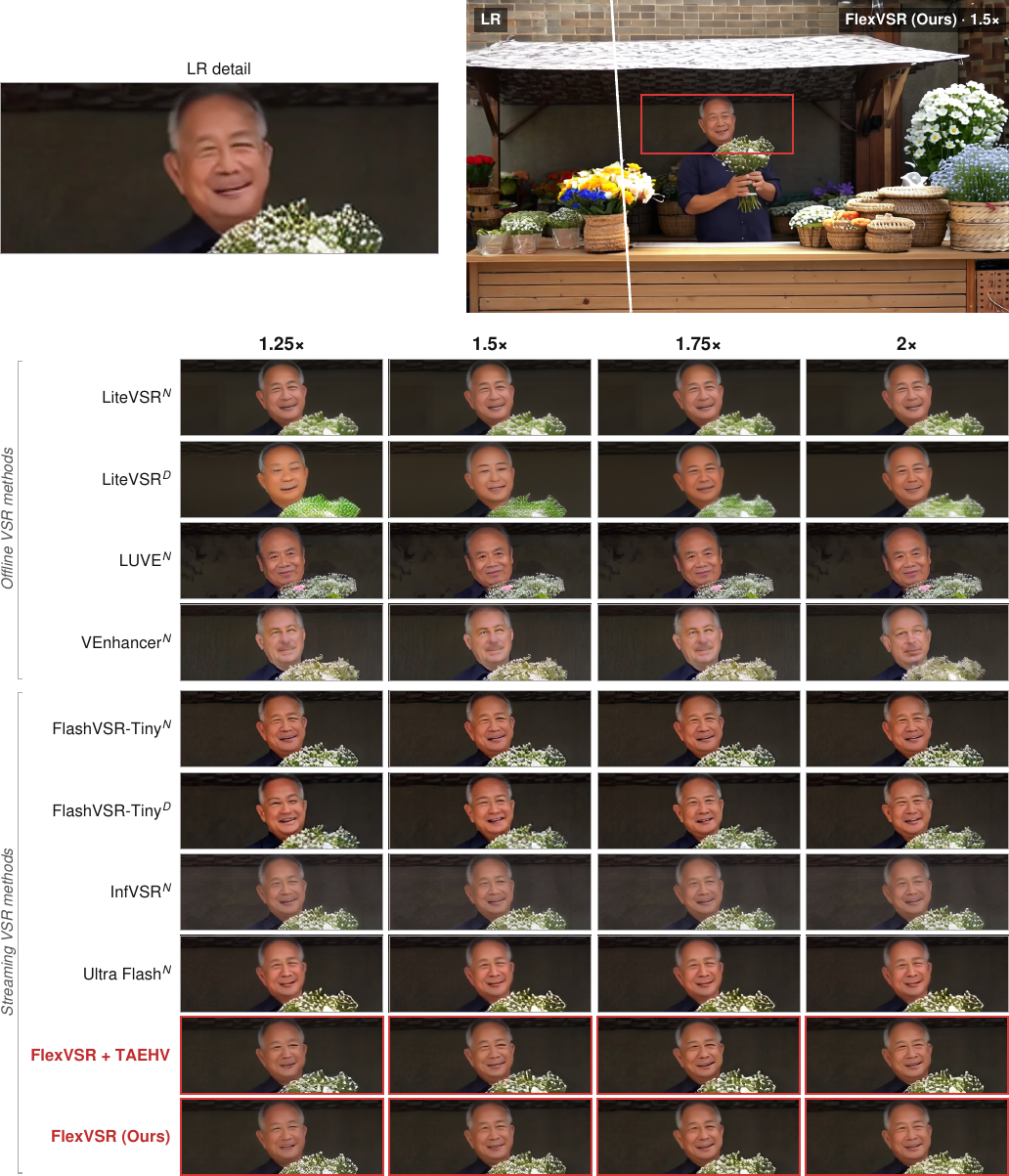}
\caption{\textbf{Continuous-scale SR comparisons.} All methods share the same upstream Wan2.1-1.3B-generated LR latent.}
\label{fig:supp_qualitative_validation_one}
\end{figure}
\clearpage
\section{Complementary Fixed-LR Evaluation}
\label{sec:supp_fixed_lr}
We evaluate the same final checkpoints on YouHQ40 while keeping each
video's LR input fixed and changing only the requested output grid.
The fixed LR sizes are $528\times960$ for 37 videos and $528\times528$
for three square videos (height $\times$ width).
For nominal scales $s\in\{1.5,1.8,2.0\}$, each target dimension is
rounded from $s$ times the corresponding LR dimension to a multiple of 16.
Each reference is resized directly from the original HR video to that target;
the $2.0\times$ reference is unchanged.
Targets are evaluated in separate runs with the same LR input and weights.

Table~\ref{tab:supp_fixed_lr} complements the fixed-HR protocol of
main-paper Table~\ref{tab:quality_comparison}. N/D processing follows Sec.~\ref{sec:supp_routes};
native outputs are resized directly to each target.
All metrics use full-frame post-SR RGB, averaged over real frames per
video and then equally over all 40 videos. FlexVSR uses the final
REDS+VSR-17K checkpoint described in Sec.~\ref{sec:supp_decoder}.

\begin{table}[!htbp]
\centering
\setlength{\aboverulesep}{0pt}\setlength{\belowrulesep}{0pt}
\caption{\textbf{Fixed-LR results on YouHQ40.} Best and second-best scores are \textbf{bold} and \underline{underlined}.}
\label{tab:supp_fixed_lr}
\fontsize{9}{11.0}\selectfont\setlength{\tabcolsep}{3pt}\renewcommand{\arraystretch}{1.05}
\sbox0{%
\begin{tabular}{@{\extracolsep{0pt}}l*{10}{c}@{}}
\toprule
\multicolumn{1}{@{}c}{\multirow{2}{*}{Method}} & \multicolumn{3}{c}{$1.5\times$} & \multicolumn{3}{c}{$1.8\times$} & \multicolumn{3}{c}{$2.0\times$} & \multirow{2}{*}{\shortstack{Avg.\\Rank$\downarrow$}} \\
\cmidrule(lr){2-4}\cmidrule(lr){5-7}\cmidrule(lr){8-10}
 & PSNR$\uparrow$ & LPIPS$\downarrow$ & tLPIPS$\downarrow$ & PSNR$\uparrow$ & LPIPS$\downarrow$ & tLPIPS$\downarrow$ & PSNR$\uparrow$ & LPIPS$\downarrow$ & tLPIPS$\downarrow$ & \\
\midrule
\multicolumn{11}{l}{\emph{Offline VSR methods}} \\
LiteVSR$^{N}$ & \textbf{26.74} & 0.1767 & \textbf{0.00792} & \textbf{26.39} & 0.2007 & \underline{0.00965} & \underline{25.97} & 0.2200 & \underline{0.01187} & \underline{2.67} \\
LiteVSR$^{D}$ & 23.16 & 0.3411 & 0.01281 & 23.29 & 0.3439 & 0.01351 & 23.35 & 0.3392 & 0.01434 & 9.00 \\
LUVE$^{N}$ & 16.43 & 0.4298 & 0.01785 & 16.38 & 0.4427 & 0.01963 & 16.32 & 0.4540 & 0.02222 & 12.00 \\
VEnhancer$^{N}$ & 19.35 & 0.3176 & 0.01364 & 19.53 & 0.3243 & 0.01560 & 20.15 & 0.3254 & 0.01798 & 10.67 \\
\midrule
\multicolumn{11}{l}{\emph{Streaming VSR methods}} \\
FlashVSR-T$^{N}$ & \underline{26.60} & 0.1455 & 0.01356 & \underline{26.12} & 0.1699 & 0.01434 & 25.72 & 0.1851 & 0.01421 & 4.67 \\
FlashVSR-T$^{D}$ & 23.23 & 0.2083 & 0.01270 & 23.56 & 0.2119 & 0.01354 & 23.67 & 0.2206 & 0.01419 & 6.78 \\
FlashVSR-F$^{N}$ & 26.15 & 0.1533 & 0.01360 & 25.65 & 0.1777 & 0.01413 & 25.26 & 0.1919 & 0.01374 & 5.22 \\
FlashVSR-F$^{D}$ & 22.78 & 0.2106 & 0.00970 & 23.08 & 0.2197 & 0.01060 & 23.20 & 0.2264 & \textbf{0.01096} & 6.78 \\
InfVSR$^{N}$ & 25.29 & 0.2173 & 0.01307 & 25.04 & 0.2286 & 0.01375 & 24.72 & 0.2483 & 0.01581 & 7.67 \\
InfVSR$^{D}$ & 24.53 & 0.2096 & 0.01250 & 24.92 & 0.2151 & 0.01294 & 24.92 & 0.2236 & 0.01435 & 6.33 \\
Ultra Flash$^{N}$ & 24.66 & \underline{0.1346} & 0.01216 & 24.46 & \underline{0.1518} & 0.01322 & 24.14 & \underline{0.1699} & 0.01416 & 4.44 \\
\midrule
FlexVSR (Ours) & 26.41 & \textbf{0.1101} & \underline{0.00904} & 26.02 & \textbf{0.1406} & \textbf{0.00877} & \textbf{26.79} & \textbf{0.1277} & 0.01242 & \textbf{1.78} \\
\bottomrule
\end{tabular}
}
\ifdim\wd0>\linewidth
  \resizebox{\linewidth}{!}{\usebox0}%
\else
  \usebox0
\fi

\par\begin{minipage}{\linewidth}\scriptsize
Avg. Rank averages the nine scale--metric ranks at displayed precision; ties receive mean ranks.
\end{minipage}
\end{table}

\FloatBarrier
\section{Decoder Quality-Speed Trade-off, Timing Protocol, and KV-Cache Variant}
\label{sec:supp_decoder}
\paragraph{Configuration.}
FlexVSR + TAEHV replaces only the final HR decoder with the released
Wan-compatible TAEHV implementation and weights used by Ultra
Flash~\cite{BoerBohan2025TAEHV,UltraFlash_2026}; no retraining is performed.
Quality and timing use the final REDS+VSR-17K ReLSA checkpoint (step 41,000),
the frozen feature extractor and SR DiT, and noise level $0.2$.
The feature extractor and ReLSA run in FP32 and the SR DiT in BF16. TAEHV uses FP16 and
sequential decoding with causal decoder memory; the default Wan decoder
uses BF16 with VAE tiling. The SR window contains two history and four
new latent frames.

\begin{table}[!htbp]
\centering
\setlength{\aboverulesep}{0pt}\setlength{\belowrulesep}{0pt}
\caption{\textbf{Complete quality results of FlexVSR + TAEHV (Ours).}
Full-frame post-SR RGB, 40 videos per dataset and scale, using the same
LR inputs and original HR references as main-paper Table~\ref{tab:quality_comparison}.
Metrics are averaged over real frames per video, then equally over videos.}
\label{tab:taehv_quality}
\small\setlength{\tabcolsep}{3pt}
\begin{tabular*}{\linewidth}{@{\extracolsep{\fill}}lc*{6}{c}@{}}
\toprule
Dataset & Scale & PSNR$\uparrow$ & SSIM$\uparrow$ & LPIPS$\downarrow$ & DISTS$\downarrow$ & tLPIPS$\downarrow$ & tOF$\downarrow$ \\
\midrule
YouHQ40 & $1.5\times$ & 26.38 & 0.7729 & 0.1421 & 0.0819 & 0.01406 & 0.4734 \\
YouHQ40 & $2\times$ & 25.98 & 0.7551 & 0.1525 & 0.0864 & 0.01476 & 0.4945 \\
Inter4K-40 & $1.5\times$ & 25.55 & 0.7747 & 0.1220 & 0.0790 & 0.00884 & 0.2093 \\
Inter4K-40 & $2\times$ & 25.56 & 0.7726 & 0.1347 & 0.0823 & 0.00837 & 0.2190 \\
\bottomrule
\end{tabular*}
\end{table}

\paragraph{Quality--speed trade-off.}
TAEHV trades reconstruction quality for decoding speed.
On YouHQ40, PSNR changes from 27.25/26.79 to 26.38/25.98~dB
at $1.5\times$/$2\times$, and LPIPS increases from 0.1122/0.1277
to 0.1421/0.1525. On Inter4K-40, LPIPS increases from
0.1040/0.1184 to 0.1220/0.1347. The faster runtime therefore corresponds
to Table~\ref{tab:taehv_quality}'s quality, not the default decoder's.

\paragraph{Timing protocol and checkpoint correspondence.}
Main-paper Table~2 benchmarks runtime on one H100 80GB GPU; FlexVSR uses
the same final REDS+VSR-17K checkpoint and settings as
Table~\ref{tab:taehv_quality}.
The output is $704\times1280$, with $464\times848$/$352\times640$
LR inputs at nominal $1.5\times$/$2\times$. A YouHQ40 clip is extended
to 77 frames by reflected temporal repetition; $T_{13}$ measures its
13-frame prefix and $t_{16}=(T_{77}-T_{13})/4$.
After one warm-up per length, SR timings use the median of three
synchronized runs, or five for rechecked configurations; the shared LR
decoder uses five runs. We use eager execution without TF32 or compilation.

\paragraph{Common latent start and decoder accounting.}
All streaming configurations in Table~2 are timed from the same prepared
upstream Wan LR latent to the target RGB output. Latent cascades
(FlexVSR, including TAEHV, and Ultra Flash) bypass LR RGB decoding and
perform only final HR decoding. Ultra Flash's native route then resizes
the decoded HR output as specified in Sec.~\ref{sec:supp_routes}.
RGB-input restorers (FlashVSR-T/-F and InfVSR, for both N and D)
require an upstream Wan VAE LR decode. For each length $n\in\{13,77\}$,
we add the independently measured stage medians:
\[
T_n=\operatorname{median}(T_n^{\mathrm{LR\ decode}})
   +\operatorname{median}(T_n^{\mathrm{SR}}).
\]
The SR term includes restoration, final HR decoding, required resizing,
and transfers. The shared LR decode contributes approximately
$0.8434/0.4517$~s to $t_{16}$ at $1.5{\times}/2{\times}$, respectively.
Subtracting this contribution gives the SR-only $t_{16}$ for an already
available LR RGB input. These totals describe serial execution without
stage overlap; upstream generation, model loading, LR encoding, and file
I/O are excluded.
In deployment, the end-to-end streaming frame rate further depends on the
target resolution, the chunk size, and whether decoding runs asynchronously
with restoration.
Quality uses the original benchmark resolutions and real frames; the
fixed-resolution timing input is a separate workload.

\begin{figure*}[t]
\centering
\includegraphics[width=\textwidth]{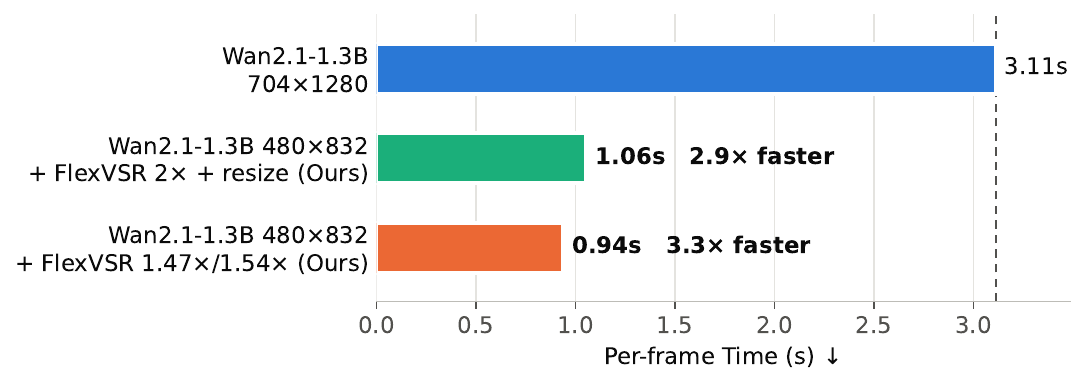}
\caption{\textbf{End-to-end per-frame latency for $704{\times}1280$ text-to-video generation} (81 frames, Wan2.1-1.3B on one H100 GPU). Generating at the official $480{\times}832$ resolution followed by FlexVSR achieves $2.9{\times}$ speedup via $2{\times}$ restoration plus RGB resizing and $3.3{\times}$ speedup via direct fractional restoration ($1.47{\times}/1.54{\times}$), both relative to direct $704{\times}1280$ sampling and producing outputs at the same resolution.}
\label{fig:supp_e2e_cascade}
\end{figure*}

\paragraph{End-to-end T2V cascade from native resolution.}
While Table~\ref{tab:efficiency} isolates restoration latency from upstream generation, Fig.~\ref{fig:supp_e2e_cascade} measures a full 81-frame $704{\times}1280$ text-to-video cascade using Wan2.1-1.3B (50-step UniPC, CFG~5.0) on one H100 GPU.
We compare direct $704{\times}1280$ sampling against two cascades that sample from the same prompt at the official $480{\times}832$ resolution and apply FlexVSR in latent space: either at $2{\times}$ ($960{\times}1664$) followed by antialiased bicubic RGB resizing, or directly at the fractional target scale ($1.47{\times}$ vertically, $1.54{\times}$ horizontally).
The amortized end-to-end latency per output frame, computed as $T_{\mathrm{total}}/81$, includes text encoding, sampling, latent restoration, and HR VAE decoding (plus RGB resizing for the $2{\times}$ route), excludes model loading, and is reported as the median over three CUDA-synchronized runs after one warm-up.
The $2{\times}$-then-resize cascade reduces latency from $3.11$ to $1.06$~s/frame ($2.9{\times}$ speedup), whereas direct fractional restoration more than halves latent restoration overhead ($0.033$ vs.\ $0.069$~s/frame) and avoids the larger $960{\times}1664$ VAE decode, reaching $0.94$~s/frame ($3.3{\times}$ speedup).

\paragraph{Historical KV-cache variant.}
FlexVSR also runs with a historical KV cache. The main-paper results
recompute history: each SR window concatenates two restored history
latent frames with four new ones, and the SR DiT processes all six.
FlexVSR-KV instead caches the per-layer attention keys and values that
the previous window computed for its last two latent frames, and runs
the SR DiT on the four new frames only. Because the history interface
changes, we retrain the SR prior and the end-to-end ReLSA stage with the
same data and budgets as the main model; the fixed-$2\times$ latent upsampler and the
latent warm-up are reused. The ReLSA architecture, the decoders, and the evaluation
protocol are unchanged.

\begin{table}[!htbp]
\centering
\caption{\textbf{Historical KV-cache variant: quality at the main-paper settings.}
Full-frame post-SR RGB, 40 videos per dataset and scale, with the LR
inputs and HR references of main-paper Table~\ref{tab:quality_comparison}. Within each decoder,
the better value is bold.}
\label{tab:supp_kv_quality}
\fontsize{8}{11.5}\selectfont      
\setlength{\tabcolsep}{0pt}        
\renewcommand{\arraystretch}{1.12} 
\setlength{\aboverulesep}{0.3ex}   
\setlength{\belowrulesep}{0.3ex}   
\begin{tabular*}{\linewidth}{@{\extracolsep{\fill}}l*{12}{c}@{}}
\toprule
\multicolumn{1}{@{}c}{\multirow{2.1}{*}{Model}} & \multicolumn{3}{c}{YouHQ40, $1.5\times$} & \multicolumn{3}{c}{YouHQ40, $2\times$}
& \multicolumn{3}{c}{Inter4K-40, $1.5\times$} & \multicolumn{3}{c}{Inter4K-40, $2\times$} \\
\cmidrule(lr){2-4}\cmidrule(lr){5-7}\cmidrule(lr){8-10}\cmidrule(lr){11-13}
 & PSNR$\uparrow$ & LPIPS$\downarrow$ & tLPIPS$\downarrow$ & PSNR$\uparrow$ & LPIPS$\downarrow$ & tLPIPS$\downarrow$
& PSNR$\uparrow$ & LPIPS$\downarrow$ & tLPIPS$\downarrow$ & PSNR$\uparrow$ & LPIPS$\downarrow$ & tLPIPS$\downarrow$ \\
\midrule
FlexVSR + Wan Decoder (main) & 27.25 & \textbf{0.1122} & 0.01244 & 26.79 & \textbf{0.1277} & 0.01242 & 26.17 & 0.1040 & \textbf{0.00723} & 26.38 & 0.1184 & \textbf{0.00801} \\
FlexVSR-KV + Wan Decoder & \textbf{27.30} & 0.1131 & \textbf{0.01178} & \textbf{26.84} & 0.1311 & \textbf{0.01187} & \textbf{26.59} & \textbf{0.0964} & 0.00764 & \textbf{26.46} & \textbf{0.1136} & 0.00838 \\
\midrule
FlexVSR + TAEHV & 26.38 & 0.1421 & \textbf{0.01406} & 25.98 & 0.1525 & \textbf{0.01476} & 25.55 & 0.1220 & \textbf{0.00884} & 25.56 & 0.1347 & \textbf{0.00837} \\
FlexVSR-KV + TAEHV & \textbf{26.39} & \textbf{0.1397} & 0.01448 & \textbf{26.04} & \textbf{0.1519} & 0.01517 & \textbf{25.91} & \textbf{0.1153} & 0.00933 & \textbf{25.85} & \textbf{0.1303} & 0.00864 \\
\bottomrule
\end{tabular*}
\end{table}

\paragraph{KV-cache quality.}
Table~\ref{tab:supp_kv_quality} compares the two models at the two
supervised scales. PSNR does not drop in any of the eight settings
($+0.01$ to $+0.42$~dB). With the default decoder, LPIPS and tLPIPS
change by small amounts in both directions: on YouHQ40, LPIPS rises by
at most 0.0034 and tLPIPS falls by about 5\%; on Inter4K-40, LPIPS
falls by 4--7\% and tLPIPS rises by about 5\%. With TAEHV, LPIPS
improves in all four settings and tLPIPS rises by 3--6\%. KV caching
thus keeps quality comparable to the main model.

\begin{table}[!htbp]
\centering
\caption{\textbf{Historical KV-cache variant: runtime} with the main-paper Table~\ref{tab:efficiency}
protocol (one H100, $704\times1280$ output, final decoding included).
$T_{13}$ and $t_{16}$ are in seconds. Within each decoder, the better
value is bold.}
\label{tab:supp_kv_runtime}
\small
\setlength{\tabcolsep}{3pt}
\renewcommand{\arraystretch}{1.12} 
\setlength{\aboverulesep}{0.3ex}   
\setlength{\belowrulesep}{0.3ex}   
\begin{tabular*}{\linewidth}{@{\extracolsep{\fill}}l*{6}{c}@{}}
\toprule
\multicolumn{1}{@{}c}{\multirow{2.1}{*}{Model}} & \multicolumn{3}{c}{$1.5\times$} & \multicolumn{3}{c}{$2\times$} \\
\cmidrule(lr){2-4}\cmidrule(lr){5-7}
 & $T_{13}\downarrow$ & $t_{16}\downarrow$ & FPS$\uparrow$ & $T_{13}\downarrow$ & $t_{16}\downarrow$ & FPS$\uparrow$ \\
\midrule
FlexVSR + Wan Decoder (main) & 1.957 & 2.387 & 6.70 & \textbf{1.885} & 2.388 & 6.70 \\
FlexVSR-KV + Wan Decoder & \textbf{1.911} & \textbf{2.287} & \textbf{7.00} & 1.893 & \textbf{2.263} & \textbf{7.07} \\
\midrule
FlexVSR + TAEHV & \textbf{0.389} & 0.674 & 23.73 & \textbf{0.385} & 0.668 & 23.95 \\
FlexVSR-KV + TAEHV & 0.407 & \textbf{0.562} & \textbf{28.44} & 0.403 & \textbf{0.558} & \textbf{28.69} \\
\bottomrule
\end{tabular*}
\par\vspace{2pt}%
\begin{minipage}{\linewidth}\scriptsize
Main-model rows repeat main-paper Table~\ref{tab:efficiency}. FlexVSR-KV is timed with the same protocol and input.
\end{minipage}
\end{table}

\paragraph{KV-cache runtime.}
Table~\ref{tab:supp_kv_runtime} follows the timing protocol above.
The cache removes the two history frames from the SR DiT input, so SR DiT's steady-state time per 16 frames drops
from 0.51 to 0.39~s; feature extraction, ReLSA, and decoding are unchanged. With the
default Wan decoder, decoding dominates and FPS rises from 6.70/6.70 to
7.00/7.07 at $1.5\times$/$2\times$. With TAEHV, FPS rises from
23.73/23.95 to 28.44/28.69. $T_{13}$ changes little because the first
chunk has no history.

\FloatBarrier
\section{Extension to a Wider Scale Range}
\label{sec:supp_wider_scales}
A separate shared ReLSA adapter extends supervision from the main model's
$1.5{\times}/2{\times}$ to $1.2\times$--$4\times$.
We sample equally from $[1.2,2)$, $[2,3)$, and $[3,4]$, using seven
training factors per interval:
\begin{align*}
\mathcal{S}_1&=\{1.2,1.3,1.4,1.5,1.6,1.8,1.9\},\\
\mathcal{S}_2&=\{2,2.2,2.3,2.4,2.5,2.7,2.9\},\\
\mathcal{S}_3&=\{3,3.2,3.3,3.4,3.5,3.8,4\}.
\end{align*}
Each factor receives $1/21$ of the supervision budget, giving 426,608
sampled training instances per interval and 1,279,824 in total.
Latent warm-up is followed by E2E training with \textbf{the same Stage-1 $2\times$ prior as the main model; only ReLSA is updated, while the feature extractor, SR DiT (trained solely on $2\times$ data), and Wan VAE remain frozen.}
One final checkpoint is used at all evaluation scales without scale-specific
fine-tuning. The five test factors $1.25\times$, $1.7\times$, $1.75\times$,
$2.75\times$, and $3.25\times$ are absent from the training set.

Table~\ref{tab:supp_wider_scales} compares this model against FlashVSR-T$^{D}$ and Ultra Flash$^{N}$, the two fastest low-latency streaming baseline configurations in main-paper Table~\ref{tab:efficiency}. All results use the same 40 YouHQ40 videos, original HR targets, and full-frame post-SR RGB metrics. FlashVSR-T (Tiny) uses its TCDecoder and full-canvas inference without additional spatial DiT tiling. Despite keeping the $2\times$-only SR DiT frozen, the wider-scale FlexVSR model achieves the best overall average rank (1.16) and leads all three metrics across all 14 test scales from $1.2\times$ through $3\times$. Against the native-$4\times$ FlashVSR-T$^{D}$, it improves PSNR by 0.17--3.60~dB at every tested factor through $3.5\times$, while FlashVSR-T$^{D}$ regains lower LPIPS and tLPIPS from $3.25\times$ onward and higher PSNR at its native $4\times$ scale. Against Ultra Flash$^{N}$, FlexVSR is better on all three metrics at every tested factor except tLPIPS at $4\times$.
\textbf{Overall, serving continuous scales with a single lightweight adapter on a frozen prior provides a practical balance among restoration quality, target-grid efficiency, and deployment flexibility for streaming video generation.}

\begingroup
\edef\SuppWideTableBase{\number\value{table}}
\setcounter{table}{\SuppWideTableBase}
\renewcommand{\theHtable}{flexsupp.wide.overall}
\begin{table}[!htbp]
\centering
\setlength{\aboverulesep}{0pt}\setlength{\belowrulesep}{0pt}
\caption{\textbf{Wider-scale comparison with the two fastest streaming baseline configurations on YouHQ40:} mean metric ranks over 17 scales ($1.2\times$--$4\times$), each with 40 videos and full-frame post-SR RGB. The overall rank weights all 51 scale--metric columns equally. Lower ranks are better; best and second-best values are \textbf{bold} and \underline{underlined}. $\dagger$ marks scales unseen during wider-scale training in the following panels.}
\label{tab:supp_wider_scales}
\fontsize{9}{11.0}\selectfont\setlength{\tabcolsep}{2.5pt}\renewcommand{\arraystretch}{1.0}
\begin{tabular*}{0.9\linewidth}{@{\extracolsep{\fill}}l*{5}{c}@{}}
\toprule
\multicolumn{1}{@{}c}{\multirow{2}{*}{Method}} & \multicolumn{3}{c}{Mean metric rank$\downarrow$} & \multirow{2}{*}{Overall Avg. Rank$\downarrow$} & \multirow{2}{*}{Position} \\
\cmidrule(lr){2-4}
 & PSNR & LPIPS & tLPIPS & & \\
\midrule
FlashVSR-T$^{D}$ & 2.59 & 2.47 & 2.47 & 2.51 & 3 \\
Ultra Flash$^{N}$ & \underline{2.35} & \underline{2.35} & \underline{2.29} & \underline{2.33} & 2 \\
\midrule
FlexVSR (Ours) & \textbf{1.06} & \textbf{1.18} & \textbf{1.24} & \textbf{1.16} & 1 \\
\bottomrule
\end{tabular*}
\par\begin{minipage}{\linewidth}\scriptsize
Ranks are computed among the three listed configurations, use the displayed precision, and average ties. Per-scale Rank averages the three metric ranks.
$N$: official native-scale inference followed by pixel-space resizing to the target size; $D$: direct target-scale inference.
Ultra Flash$^{N}$ uses its $2\times$ models up to $2\times$ and its $3\times$ models above (Sec.~\ref{sec:supp_routes}).
-T denotes the Tiny decoder of FlashVSR.
The same original HR reference is used for every method.
\end{minipage}
\end{table}

\setcounter{table}{\SuppWideTableBase}
\renewcommand{\theHtable}{flexsupp.wide.1}
\begin{table}[!htbp]
\centering
\setlength{\aboverulesep}{0pt}\setlength{\belowrulesep}{0pt}
\caption{Continued: quality at $1.2\times$, $1.25\times$$^{\dagger}$, $1.3\times$.}
\fontsize{9}{11.0}\selectfont\setlength{\tabcolsep}{2.5pt}\renewcommand{\arraystretch}{1.0}

\sbox0{%
\begin{tabular}{@{\extracolsep{0pt}}l*{12}{c}@{}}
\toprule
\multicolumn{1}{@{}c}{\multirow{2}{*}{Method}} & \multicolumn{4}{c}{$1.2\times$} & \multicolumn{4}{c}{$1.25\times$$^{\dagger}$} & \multicolumn{4}{c}{$1.3\times$} \\
\cmidrule(lr){2-5}\cmidrule(lr){6-9}\cmidrule(lr){10-13}
 & PSNR$\uparrow$ & LPIPS$\downarrow$ & tLPIPS$\downarrow$ & Rank$\downarrow$ & PSNR$\uparrow$ & LPIPS$\downarrow$ & tLPIPS$\downarrow$ & Rank$\downarrow$ & PSNR$\uparrow$ & LPIPS$\downarrow$ & tLPIPS$\downarrow$ & Rank$\downarrow$ \\
\midrule
FlashVSR-T$^{D}$ & 23.64 & 0.2209 & 0.01430 & 3.00 & 23.64 & 0.2209 & 0.01429 & 3.00 & 23.64 & 0.2208 & 0.01428 & 3.00 \\
Ultra Flash$^{N}$ & \underline{25.80} & \underline{0.1225} & \underline{0.01199} & \underline{2.00} & \underline{25.65} & \underline{0.1258} & \underline{0.01220} & \underline{2.00} & \underline{25.54} & \underline{0.1285} & \underline{0.01234} & \underline{2.00} \\
\midrule
FlexVSR (Ours) & \textbf{27.08} & \textbf{0.1116} & \textbf{0.01161} & \textbf{1.00} & \textbf{27.16} & \textbf{0.1105} & \textbf{0.01170} & \textbf{1.00} & \textbf{27.22} & \textbf{0.1101} & \textbf{0.01170} & \textbf{1.00} \\
\bottomrule
\end{tabular}
}
\ifdim\wd0>\linewidth
  \resizebox{\linewidth}{!}{\usebox0}%
\else
  \usebox0
\fi

\end{table}

\setcounter{table}{\SuppWideTableBase}
\renewcommand{\theHtable}{flexsupp.wide.2}
\begin{table}[!htbp]
\centering
\setlength{\aboverulesep}{0pt}\setlength{\belowrulesep}{0pt}
\caption{Continued: quality at $1.4\times$, $1.5\times$, $1.6\times$.}
\fontsize{9}{11.0}\selectfont\setlength{\tabcolsep}{2.5pt}\renewcommand{\arraystretch}{1.0}

\sbox0{%
\begin{tabular}{@{\extracolsep{0pt}}l*{12}{c}@{}}
\toprule
\multicolumn{1}{@{}c}{\multirow{2}{*}{Method}} & \multicolumn{4}{c}{$1.4\times$} & \multicolumn{4}{c}{$1.5\times$} & \multicolumn{4}{c}{$1.6\times$} \\
\cmidrule(lr){2-5}\cmidrule(lr){6-9}\cmidrule(lr){10-13}
 & PSNR$\uparrow$ & LPIPS$\downarrow$ & tLPIPS$\downarrow$ & Rank$\downarrow$ & PSNR$\uparrow$ & LPIPS$\downarrow$ & tLPIPS$\downarrow$ & Rank$\downarrow$ & PSNR$\uparrow$ & LPIPS$\downarrow$ & tLPIPS$\downarrow$ & Rank$\downarrow$ \\
\midrule
FlashVSR-T$^{D}$ & 23.65 & 0.2208 & 0.01427 & 3.00 & 23.65 & 0.2209 & 0.01426 & 3.00 & 23.66 & 0.2207 & 0.01425 & 3.00 \\
Ultra Flash$^{N}$ & \underline{25.29} & \underline{0.1349} & \underline{0.01267} & \underline{2.00} & \underline{25.10} & \underline{0.1405} & \underline{0.01298} & \underline{2.00} & \underline{24.85} & \underline{0.1477} & \underline{0.01328} & \underline{2.00} \\
\midrule
FlexVSR (Ours) & \textbf{27.25} & \textbf{0.1109} & \textbf{0.01184} & \textbf{1.00} & \textbf{27.23} & \textbf{0.1118} & \textbf{0.01257} & \textbf{1.00} & \textbf{27.11} & \textbf{0.1161} & \textbf{0.01206} & \textbf{1.00} \\
\bottomrule
\end{tabular}
}
\ifdim\wd0>\linewidth
  \resizebox{\linewidth}{!}{\usebox0}%
\else
  \usebox0
\fi

\end{table}

\setcounter{table}{\SuppWideTableBase}
\renewcommand{\theHtable}{flexsupp.wide.3}
\begin{table}[!htbp]
\centering
\setlength{\aboverulesep}{0pt}\setlength{\belowrulesep}{0pt}
\caption{Continued: quality at $1.7\times$$^{\dagger}$, $1.75\times$$^{\dagger}$, $1.8\times$.}
\fontsize{9}{11.0}\selectfont\setlength{\tabcolsep}{2.5pt}\renewcommand{\arraystretch}{1.0}

\sbox0{%
\begin{tabular}{@{\extracolsep{0pt}}l*{12}{c}@{}}
\toprule
\multicolumn{1}{@{}c}{\multirow{2}{*}{Method}} & \multicolumn{4}{c}{$1.7\times$$^{\dagger}$} & \multicolumn{4}{c}{$1.75\times$$^{\dagger}$} & \multicolumn{4}{c}{$1.8\times$} \\
\cmidrule(lr){2-5}\cmidrule(lr){6-9}\cmidrule(lr){10-13}
 & PSNR$\uparrow$ & LPIPS$\downarrow$ & tLPIPS$\downarrow$ & Rank$\downarrow$ & PSNR$\uparrow$ & LPIPS$\downarrow$ & tLPIPS$\downarrow$ & Rank$\downarrow$ & PSNR$\uparrow$ & LPIPS$\downarrow$ & tLPIPS$\downarrow$ & Rank$\downarrow$ \\
\midrule
FlashVSR-T$^{D}$ & 23.66 & 0.2207 & 0.01425 & 3.00 & 23.66 & 0.2207 & 0.01423 & 3.00 & 23.66 & 0.2207 & 0.01423 & 3.00 \\
Ultra Flash$^{N}$ & \underline{24.71} & \underline{0.1528} & \underline{0.01341} & \underline{2.00} & \underline{24.60} & \underline{0.1565} & \underline{0.01358} & \underline{2.00} & \underline{24.52} & \underline{0.1596} & \underline{0.01371} & \underline{2.00} \\
\midrule
FlexVSR (Ours) & \textbf{26.99} & \textbf{0.1200} & \textbf{0.01209} & \textbf{1.00} & \textbf{26.91} & \textbf{0.1226} & \textbf{0.01212} & \textbf{1.00} & \textbf{26.83} & \textbf{0.1256} & \textbf{0.01215} & \textbf{1.00} \\
\bottomrule
\end{tabular}
}
\ifdim\wd0>\linewidth
  \resizebox{\linewidth}{!}{\usebox0}%
\else
  \usebox0
\fi

\end{table}

\setcounter{table}{\SuppWideTableBase}
\renewcommand{\theHtable}{flexsupp.wide.4}
\begin{table}[!htbp]
\centering
\setlength{\aboverulesep}{0pt}\setlength{\belowrulesep}{0pt}
\caption{Continued: quality at $1.9\times$, $2\times$, $2.5\times$.}
\fontsize{9}{11.0}\selectfont\setlength{\tabcolsep}{2.5pt}\renewcommand{\arraystretch}{1.0}

\sbox0{%
\begin{tabular}{@{\extracolsep{0pt}}l*{12}{c}@{}}
\toprule
\multicolumn{1}{@{}c}{\multirow{2}{*}{Method}} & \multicolumn{4}{c}{$1.9\times$} & \multicolumn{4}{c}{$2\times$} & \multicolumn{4}{c}{$2.5\times$} \\
\cmidrule(lr){2-5}\cmidrule(lr){6-9}\cmidrule(lr){10-13}
 & PSNR$\uparrow$ & LPIPS$\downarrow$ & tLPIPS$\downarrow$ & Rank$\downarrow$ & PSNR$\uparrow$ & LPIPS$\downarrow$ & tLPIPS$\downarrow$ & Rank$\downarrow$ & PSNR$\uparrow$ & LPIPS$\downarrow$ & tLPIPS$\downarrow$ & Rank$\downarrow$ \\
\midrule
FlashVSR-T$^{D}$ & 23.67 & 0.2206 & 0.01421 & 3.00 & 23.67 & 0.2206 & 0.01419 & 3.00 & \underline{23.69} & \underline{0.2207} & \underline{0.01411} & \underline{2.00} \\
Ultra Flash$^{N}$ & \underline{24.36} & \underline{0.1660} & \underline{0.01388} & \underline{2.00} & \underline{24.14} & \underline{0.1699} & \underline{0.01416} & \underline{2.00} & 18.44 & 0.2466 & 0.01459 & 3.00 \\
\midrule
FlexVSR (Ours) & \textbf{26.62} & \textbf{0.1329} & \textbf{0.01221} & \textbf{1.00} & \textbf{26.64} & \textbf{0.1378} & \textbf{0.01331} & \textbf{1.00} & \textbf{25.44} & \textbf{0.1811} & \textbf{0.01310} & \textbf{1.00} \\
\bottomrule
\end{tabular}
}
\ifdim\wd0>\linewidth
  \resizebox{\linewidth}{!}{\usebox0}%
\else
  \usebox0
\fi

\end{table}

\setcounter{table}{\SuppWideTableBase}
\renewcommand{\theHtable}{flexsupp.wide.5}
\begin{table}[!htbp]
\centering
\setlength{\aboverulesep}{0pt}\setlength{\belowrulesep}{0pt}
\caption{Continued: quality at $2.75\times$$^{\dagger}$, $3\times$, $3.25\times$$^{\dagger}$.}
\fontsize{9}{11.0}\selectfont\setlength{\tabcolsep}{2.5pt}\renewcommand{\arraystretch}{1.0}

\sbox0{%
\begin{tabular}{@{\extracolsep{0pt}}l*{12}{c}@{}}
\toprule
\multicolumn{1}{@{}c}{\multirow{2}{*}{Method}} & \multicolumn{4}{c}{$2.75\times$$^{\dagger}$} & \multicolumn{4}{c}{$3\times$} & \multicolumn{4}{c}{$3.25\times$$^{\dagger}$} \\
\cmidrule(lr){2-5}\cmidrule(lr){6-9}\cmidrule(lr){10-13}
 & PSNR$\uparrow$ & LPIPS$\downarrow$ & tLPIPS$\downarrow$ & Rank$\downarrow$ & PSNR$\uparrow$ & LPIPS$\downarrow$ & tLPIPS$\downarrow$ & Rank$\downarrow$ & PSNR$\uparrow$ & LPIPS$\downarrow$ & tLPIPS$\downarrow$ & Rank$\downarrow$ \\
\midrule
FlashVSR-T$^{D}$ & \underline{23.69} & \underline{0.2208} & \underline{0.01406} & \underline{2.00} & \underline{23.69} & \underline{0.2209} & \underline{0.01400} & \underline{2.00} & \underline{23.69} & \textbf{0.2214} & \textbf{0.01397} & \textbf{1.33} \\
Ultra Flash$^{N}$ & 18.38 & 0.2590 & 0.01488 & 3.00 & 18.32 & 0.2702 & 0.01528 & 3.00 & 18.26 & 0.2920 & 0.01529 & 3.00 \\
\midrule
FlexVSR (Ours) & \textbf{25.06} & \textbf{0.1997} & \textbf{0.01347} & \textbf{1.00} & \textbf{24.64} & \textbf{0.2190} & \textbf{0.01383} & \textbf{1.00} & \textbf{24.17} & \underline{0.2439} & \underline{0.01441} & \underline{1.67} \\
\bottomrule
\end{tabular}
}
\ifdim\wd0>\linewidth
  \resizebox{\linewidth}{!}{\usebox0}%
\else
  \usebox0
\fi

\end{table}

\setcounter{table}{\SuppWideTableBase}
\renewcommand{\theHtable}{flexsupp.wide.6}
\begin{table}[!htbp]
\centering
\setlength{\aboverulesep}{0pt}\setlength{\belowrulesep}{0pt}
\caption{Continued: quality at $3.5\times$, $4\times$.}
\fontsize{9}{11.0}\selectfont\setlength{\tabcolsep}{2.5pt}\renewcommand{\arraystretch}{1.0}

\sbox0{%
\begin{tabular}{@{}l*{8}{c}@{}}
\toprule
\multicolumn{1}{@{}c}{\multirow{2}{*}{Method}} & \multicolumn{4}{c}{$3.5\times$} & \multicolumn{4}{c}{$4\times^{\ddagger}$} \\
\cmidrule(lr){2-5}\cmidrule(lr){6-9}
 & PSNR$\uparrow$ & LPIPS$\downarrow$ & tLPIPS$\downarrow$ & Rank$\downarrow$ & PSNR$\uparrow$ & LPIPS$\downarrow$ & tLPIPS$\downarrow$ & Rank$\downarrow$ \\
\midrule
FlashVSR-T$^{D}$ & \underline{23.68} & \textbf{0.2219} & \textbf{0.01394} & \textbf{1.33} & \textbf{23.65} & \textbf{0.2236} & \textbf{0.01387} & \textbf{1.00} \\
Ultra Flash$^{N}$ & 18.22 & 0.3061 & 0.01573 & 3.00 & 18.17 & 0.3360 & \underline{0.01589} & 2.67 \\
\midrule
FlexVSR (Ours) & \textbf{23.85} & \underline{0.2617} & \underline{0.01499} & \underline{1.67} & \underline{23.08} & \underline{0.3097} & 0.01597 & \underline{2.33} \\
\bottomrule
\end{tabular}
}
\ifdim\wd0>\linewidth
  \resizebox{\linewidth}{!}{\usebox0}%
\else
  \usebox0
\fi

\par\vspace{2pt}
\noindent{\fontsize{9}{11.0}\selectfont $^{\ddagger}$At $4\times$, FlashVSR-T$^{D}$ coincides with its native $4\times$ inference.}
\end{table}
\endgroup

\FloatBarrier
\section{Native and Direct Route Specifications}
\label{sec:supp_routes}
Let $X_L\in\mathbb{R}^{T\times3\times H_L\times W_L}$ be the common
LR RGB input and $(H_t,W_t)$ the target size.
$N$ applies native inference followed by target-size RGB resizing;
$D$ restores on the target grid via input resizing, without retraining.
All four D configurations in Table~\ref{tab:direct_routes} upsample RGB
directly to $(H_t,W_t)$ \emph{before} feature or latent encoding:
there is no preliminary LR downsampling to $(H_t/4,W_t/4)$ or
interpolation of encoded latents.

\begin{table}[!htbp]
\centering
\setlength{\aboverulesep}{0pt}\setlength{\belowrulesep}{0pt}
\caption{\textbf{Exact input processing for D routes.} Sizes are height $\times$
width. All start from the common LR RGB frames, quantized to uint8 as
expected by the released video-input pipelines.}
\label{tab:direct_routes}
\small\setlength{\tabcolsep}{4pt}\renewcommand{\arraystretch}{1.15}
\begin{tabular}{@{}>{\raggedright\arraybackslash}m{0.16\linewidth}>{\raggedright\arraybackslash}m{0.37\linewidth}>{\raggedright\arraybackslash}m{0.42\linewidth}@{}}
\toprule
Method & RGB resize before the model & Subsequent processing \\
\midrule
FlashVSR-T$^{D}$~\cite{Zhuang_2026_CVPR}
& PIL bicubic to $(H_t,W_t)$; normalize to $[-1,1]$ in BF16.
& Reflect-pad right/bottom to multiples of 128; LQ projection, DiT, and TCDecoder; remove padding. \\
FlashVSR-F$^{D}$~\cite{Zhuang_2026_CVPR}
& Same RGB resize and padding as FlashVSR-T$^{D}$.
& Same target-grid LQ conditioning and DiT; Wan VAE HR decoder replaces TCDecoder. \\
InfVSR$^{D}$~\cite{pmlr-v306-zhang26dm}
& Reshape each RGB chunk from $BCTHW$ to $(BT)CHW$; FP32 bicubic with antialiasing to $(H_t,W_t)$.
& Restore the temporal layout, then Wan VAE encoding, causal DiT restoration, and Wan VAE decoding. \\
LiteVSR$^{D}$~\cite{pmlr-v306-cao26n}
& FP32 bilinear resize of the $TCHW$ RGB tensor to $(H_t,W_t)$; \texttt{align\_corners=False}; normalize to $[-1,1]$.
& Wan2.2 VAE encoding, five-step UniPC restoration with the released adapter, and VAE decoding. \\
\bottomrule
\end{tabular}
\end{table}

For InfVSR and LiteVSR, the evaluation wrapper supplies an explicit
target size in place of the integer-scale resize helper; weights are
unchanged. LiteVSR retains the official 21-frame chunks with eight-frame
overlap and, above $1024^2$ output pixels, $704\times1280$ spatial tiles
(orientation-adjusted) with 32-pixel overlap. FlashVSR uses the full
spatial canvas without additional DiT spatial tiling; padding preserves
the benchmark field of view. Tiny/Full identifies the HR decoder,
independently of N/D routing.

\paragraph{Native routes and other methods.}
For FlashVSR-T/-F, InfVSR, and LiteVSR, N restores at
$(4H_L,4W_L)$ using the corresponding RGB interpolation above, then
applies antialiased bicubic RGB resizing to $(H_t,W_t)$.
The other baselines have only N entries: Ultra Flash uses its released
$2\times$ (1K) latent upsampler and SR DiT for all targets up to $2\times$,
LUVE~\cite{pmlr-v306-zhao26d} uses its native 2K latent target grid, and VEnhancer~\cite{he2024venhancer} uses its
official resolution-adjustment rule. Their RGB outputs are matched to
the target size. Thus N refers to each method's native inference setting,
not a universal $4\times$ setting.
For targets above $2\times$ (Table~\ref{tab:supp_extrapolation} and Table~\ref{tab:supp_wider_scales}),
Ultra Flash$^{N}$ switches to its released $3\times$ (2K) latent upsampler and SR DiT,
whose RGB output is downsampled for targets up to $3\times$ and upsampled beyond.
Under the released inference code, the outputs of these $3\times$ models are spatially displaced from
the HR reference (median 9.9~px at $3\times$ by block-wise phase correlation),
which mainly penalizes pixel-aligned PSNR;
as a control, the $2\times$ models applied to the same $2.5\times$ inputs with
RGB upsampling show no such displacement (1.0~px) and reach 23.57~dB,
still below FlexVSR (25.44~dB).

\paragraph{System-level routing vs.\ within-backbone ablations.}
Across main-paper Tables~1--3, N and D represent two practical inference
routes for producing the requested output resolution using fixed-scale
baselines. Because these routes differ in input preprocessing, conditioning,
restoration grid, and computational and decoding workloads, their comparisons
evaluate end-to-end system-level performance. Complementing these system-level
evaluations, the within-backbone ablations in main-paper Table~\ref{tab:ablation_refinement} share the
same YouHQ40 inputs and HR references while keeping the Stage-1 SR prior
frozen, providing more targeted evidence for the learned target-grid latent
interface.

\FloatBarrier
\section{Complete Metrics on YouHQ40}
\label{sec:supp_full_metrics}
Table~\ref{tab:supp_full_metrics} reports six metrics for the twelve
configurations of main-paper Table~\ref{tab:quality_comparison} on YouHQ40, using the same outputs.
PSNR, LPIPS, and tLPIPS~\cite{chu2020tecogan} repeat Table~1; SSIM~\cite{wang2004ssim}, DISTS, and tOF~\cite{chu2020tecogan} are added.
tOF is the mean absolute difference between the Farneb\"ack optical flow~\cite{Farneback_2003_Flow}
of adjacent output frames and that of the reference frames.
FlexVSR has the best PSNR, LPIPS, and DISTS at both scales.
The native routes of LiteVSR and FlashVSR have higher SSIM, and
LiteVSR$^{N}$ has the lowest tOF; FlexVSR ranks fourth and second in
tOF at $1.5\times$ and $2\times$.

\begin{table}[!htbp]
\centering
\setlength{\aboverulesep}{0pt}\setlength{\belowrulesep}{0pt}
\caption{\textbf{Six metrics on YouHQ40} for the configurations of main-paper Table~\ref{tab:quality_comparison}.
Full-frame post-SR RGB against original HR, 40 videos per scale.
Best and second-best scores are \textbf{bold} and \underline{underlined}.}
\label{tab:supp_full_metrics}
\fontsize{9}{11.0}\selectfont\setlength{\tabcolsep}{3pt}\renewcommand{\arraystretch}{1.05}
\sbox0{%
\begin{tabular}{@{\extracolsep{0pt}}l*{12}{c}@{}}
\toprule
\multicolumn{1}{@{}c}{\multirow{2}{*}{Method}} & \multicolumn{6}{c}{$1.5\times$} & \multicolumn{6}{c}{$2\times$} \\
\cmidrule(lr){2-7}\cmidrule(lr){8-13}
 & PSNR$\uparrow$ & SSIM$\uparrow$ & LPIPS$\downarrow$ & DISTS$\downarrow$ & tLPIPS$\downarrow$ & tOF$\downarrow$ & PSNR$\uparrow$ & SSIM$\uparrow$ & LPIPS$\downarrow$ & DISTS$\downarrow$ & tLPIPS$\downarrow$ & tOF$\downarrow$ \\
\midrule
\multicolumn{13}{l}{\emph{Offline VSR methods}} \\
LiteVSR$^{N}$ & \underline{26.79} & \textbf{0.8240} & 0.1943 & 0.1172 & \textbf{0.01038} & \textbf{0.3742} & \underline{25.97} & \underline{0.7805} & 0.2200 & 0.1279 & \underline{0.01187} & \textbf{0.4115} \\
LiteVSR$^{D}$ & 23.06 & 0.6179 & 0.3576 & 0.1784 & 0.01470 & 0.5923 & 23.35 & 0.6334 & 0.3392 & 0.1712 & 0.01434 & 0.5772 \\
LUVE$^{N}$ & 16.64 & 0.4558 & 0.4398 & 0.1571 & 0.02030 & 0.9688 & 16.32 & 0.4302 & 0.4540 & 0.1651 & 0.02222 & 0.9889 \\
VEnhancer$^{N}$ & 19.90 & 0.5406 & 0.3274 & 0.1395 & 0.01736 & 0.7017 & 20.15 & 0.5525 & 0.3254 & 0.1377 & 0.01798 & 0.6844 \\
\midrule
\multicolumn{13}{l}{\emph{Streaming VSR methods}} \\
FlashVSR-T$^{N}$ & 26.51 & \underline{0.8191} & 0.1723 & 0.0944 & 0.01440 & \underline{0.3840} & 25.72 & \textbf{0.7827} & 0.1851 & 0.0975 & 0.01421 & 0.4392 \\
FlashVSR-T$^{D}$ & 23.65 & 0.6686 & 0.2209 & 0.1043 & 0.01426 & 0.6010 & 23.67 & 0.6691 & 0.2206 & 0.1042 & 0.01419 & 0.5998 \\
FlashVSR-F$^{N}$ & 26.04 & 0.8106 & 0.1805 & 0.0981 & 0.01456 & 0.3889 & 25.26 & 0.7732 & 0.1919 & 0.1013 & 0.01374 & 0.4396 \\
FlashVSR-F$^{D}$ & 23.17 & 0.6558 & 0.2267 & 0.1078 & \underline{0.01101} & 0.5682 & 23.20 & 0.6563 & 0.2264 & 0.1076 & \textbf{0.01096} & 0.5673 \\
InfVSR$^{N}$ & 24.47 & 0.7456 & 0.2981 & 0.1745 & 0.01896 & 0.5020 & 24.72 & 0.7313 & 0.2483 & 0.1369 & 0.01581 & 0.5451 \\
InfVSR$^{D}$ & 24.72 & 0.7019 & 0.2289 & 0.1105 & 0.01400 & 0.6031 & 24.92 & 0.7065 & 0.2236 & 0.1056 & 0.01435 & 0.6016 \\
Ultra Flash$^{N}$ & 25.10 & 0.7722 & \underline{0.1405} & \underline{0.0839} & 0.01298 & 0.4580 & 24.14 & 0.7162 & \underline{0.1699} & \underline{0.0942} & 0.01416 & 0.5490 \\
\midrule
FlexVSR (Ours) & \textbf{27.25} & 0.7875 & \textbf{0.1122} & \textbf{0.0755} & 0.01244 & 0.3947 & \textbf{26.79} & 0.7671 & \textbf{0.1277} & \textbf{0.0922} & 0.01242 & \underline{0.4144} \\
\bottomrule
\end{tabular}
}
\ifdim\wd0>\linewidth
  \resizebox{\linewidth}{!}{\usebox0}%
\else
  \usebox0
\fi

\end{table}

\FloatBarrier
\Needspace{16\baselineskip}
\section{Above-Range Extrapolation}
\label{sec:supp_extrapolation}

Table~\ref{tab:supp_extrapolation} reports the detailed YouHQ40 scores at $2.1\times$ and $2.2\times$ referenced in Sec.~\ref{sec:scale_generalization}. Note that Ultra Flash$^N$ switches to its released $3\times$ models above $2\times$ (Sec.~\ref{sec:supp_routes}).

\begin{table}[!htb]
\centering
\setlength{\aboverulesep}{0pt}\setlength{\belowrulesep}{0pt}
\caption{\textbf{Above-range extrapolation ($2.1\times$--$2.2\times$) on YouHQ40} without retraining (conventions follow Table~\ref{tab:cross_scale}).}
\label{tab:supp_extrapolation}
\fontsize{9}{11.0}\selectfont
\setlength{\tabcolsep}{0pt}
\renewcommand{\arraystretch}{1.15}
\begin{tabular*}{\linewidth}{@{}l@{\extracolsep{\fill}}*{6}{c}@{}}
\toprule
\multicolumn{1}{@{}c@{}}{\multirow{2}{*}{Method}} & \multicolumn{3}{c}{$2.1\times$} & \multicolumn{3}{c@{}}{$2.2\times$} \\
\cmidrule(lr){2-4}\cmidrule(l){5-7}
& PSNR$\uparrow$ & LPIPS$\downarrow$ & tLPIPS$\downarrow$ & PSNR$\uparrow$ & LPIPS$\downarrow$ & tLPIPS$\downarrow$ \\
\midrule
FlashVSR-T$^{N}$ & \textbf{25.52} & \underline{0.1897} & \underline{0.01404} & \textbf{25.44} & \underline{0.1911} & \underline{0.01386} \\
FlashVSR-T$^{D}$ & 23.68 & 0.2207 & 0.01416 & 23.68 & 0.2206 & 0.01416 \\
InfVSR$^{N}$ & 24.79 & 0.2471 & 0.01572 & 24.84 & 0.2462 & 0.01570 \\
InfVSR$^{D}$ & 24.93 & 0.2238 & 0.01446 & \underline{24.94} & 0.2242 & 0.01453 \\
Ultra Flash$^{N}$ & 18.72 & 0.2227 & 0.01420 & 18.63 & 0.2284 & 0.01440 \\
\midrule
FlexVSR (Ours) & \underline{24.98} & \textbf{0.1589} & \textbf{0.01135} & 24.67 & \textbf{0.1756} & \textbf{0.01194} \\
\bottomrule
\end{tabular*}
\end{table}

\end{document}